\documentclass[letterpaper]{article} %
\usepackage{aaai25}  %
\usepackage{times}  %
\usepackage{helvet}  %
\usepackage{courier}  %
\usepackage[hyphens]{url}  %
\usepackage{graphicx} %
\usepackage{natbib}  %
\usepackage{caption} %
\usepackage{algorithm}
\usepackage{algorithmic}

\usepackage{subcaption}
\usepackage{amsmath,amssymb,amsfonts} %
\usepackage{arydshln} %
\usepackage{nameref} %
\usepackage{multirow} %
\usepackage{booktabs} %
\usepackage{enumitem}

\usepackage[english]{babel} %

\usepackage{amsthm}

\newtheorem{proposition}{Proposition}

\newtheorem{definition}{Definition}

\usepackage{tikz}

\newcommand{\commentout}[1]{{}}

\newcommand{\eat}[1]{}

\usepackage{newfloat}
\usepackage{listings}
\DeclareCaptionStyle{ruled}{labelfont=normalfont,labelsep=colon,strut=off} %
\floatstyle{ruled}
\newfloat{listing}{tb}{lst}{}
\floatname{listing}{Listing}
\nocopyright

\title{\textsc{GraSP}: Simple yet Effective Graph Similarity Predictions}
\author{
    Haoran Zheng\textsuperscript{\rm 1}\thanks{Work partially done at the Hong Kong Polytechnic University.}, Jieming Shi\textsuperscript{\rm 2}\thanks{Corresponding author.}, and Renchi Yang\textsuperscript{\rm 1}
}
\affiliations{
    \textsuperscript{\rm 1}Hong Kong Baptist University, Hong Kong, China\\
    \textsuperscript{\rm 2}The Hong Kong Polytechnic University, Hong Kong, China\\

    cshrzheng@comp.hkbu.edu.hk, jieming.shi@polyu.edu.hk, renchi@hkbu.edu.hk
}

\usepackage{bibentry}

\begin{document}

\maketitle

\begin{abstract}
Graph similarity computation (GSC) is to calculate the similarity between one pair of graphs, which is a fundamental problem with fruitful applications in the graph community. In GSC, {graph edit distance} (GED) and {maximum common subgraph} (MCS) are two important similarity metrics, both of which are NP-hard to compute. Instead of calculating the exact values, recent solutions resort to leveraging {graph neural networks} (GNNs) to learn data-driven models for the estimation of GED and MCS. Most of them are built on components involving node-level interactions crossing graphs, which engender vast computation overhead but are of little avail in effectiveness.
In the paper, we present \textsc{GraSP}, a simple yet effective GSC approach for GED and MCS prediction. \textsc{GraSP} achieves high result efficacy through several key instruments: enhanced node features via positional encoding and a GNN model augmented by a gating mechanism, residual connections, as well as multi-scale pooling.
Theoretically, \textsc{GraSP} can surpass the 1-WL test, indicating its high expressiveness.
Empirically, extensive experiments comparing \textsc{GraSP} against {10} competitors on multiple widely adopted benchmark datasets showcase the superiority of \textsc{GraSP} over prior arts in terms of both effectiveness and efficiency.
The code is available at \url{https://github.com/HaoranZ99/GraSP}.
\end{abstract}

\section{Introduction} \label{section: 1}
Graphs are an omnipresent data structure to model complex relationships between objects in nature and society, such as social networks, 
transportation networks, and biological networks.
Graph similarity calculation (hereinafter GSC) is a fundamental operation for graphs and finds widespread use in applications,
e.g., ranking documents \cite{lee_rank-based_2008}, disease prediction \cite{borgwardt_graph_2006}, and code similarity detection \cite{li_graph_2019}. In GSC, {\em graph edit distance} (GED) \cite{bunke_inexact_1983} and {\em maximum common subgraph} (MCS) \cite{bunke_graph_1998} are two popular similarity metrics. %
However, the exact computations of GED and MCS are both NP-hard \cite{zeng_comparing_2009, liu_learning_2020}. As revealed in~\cite{blumenthal_exact_2020}, calculating the exact GED between graphs containing over just 16 nodes is infeasible.

Early attempts have been made to trade accuracy for efficiency via combinatorial approximation \cite{neuhaus_fast_2006,riesen_approximate_2009,fankhauser_speeding_2011}. However, such methods still incur sub-exponential or cubic asymptotic complexities, rendering them impractical for sizable graphs or a large collection of graphs.
In recent years, inspired by the powerful capabilities of {graph neural networks} (GNNs) in capturing graph structures, a series of GNN-based data-driven models \cite{bai_simgnn_2020, li_graph_2019, bai_learning-based_2020, ling_multilevel_2021,tan_exploring_2023, piao_computing_2023} have been developed for GSC prediction.
At a high level, most of them usually encompass three components, including the node/graph embedding module, the cross-graph node-level interaction module, as well as the similarity computation module. Amid them, modeling the node-level interactions that cross graphs leads to a quadratic time complexity in offline training or online inference, significantly impeding the efficiency and scalability of such models. Efforts have been made to alleviate this \cite{qin_slow_2021, zhuo_efficient_2022, ranjan_greed_2023}, but there is still significant space for improvement.

In this paper, to  achieve superior efficacy,
we present \textsc{GraSP} (short for \underline{GRA}ph \underline{S}imilarity \underline{P}rediction), a simple but effective  approach for GED/MCS prediction. Notably, \textsc{GraSP} can achieve efficient training and inference while offering superb performance in GED/MCS predictions.
The design recipes for \textsc{GraSP} include several major ingredients. \textsc{GraSP} first utilizes positional encoding to inject global topological patterns in graphs for enhanced node features. The augmented node features are then fed into a carefully crafted graph embedding architecture, which comprises an optimized GNN backbone for learning node representations
and additionally, a {multi-scale pooling} trick that capitalizes on the merits of different pooling schemes to transform the node representations
into high-caliber graph-level embeddings. 
As such, the GED/MCS prediction can be directly solved with such graph embeddings, without acquiring all pair-wise node interactions for node embeddings or cross-level similarities required in previous methods.
Furthermore, compared to conventional GNN models \cite{kipf_semi-supervised_2017,xu_how_2019} with expressive power bounded by the 1-order Weisfeiler-Lehman test (1-WL test) \cite{weisfeiler_reduction_1968}, our analysis manifests that \textsc{GraSP} obtains higher expressiveness by surpassing the 1-WL test, theoretically corroborating its practical effectiveness in accurate GSC. We experimentally evaluate \textsc{GraSP} against {10} competitors over 4 real datasets in GED and MCS prediction tasks under various settings. The empirical results exhibit that \textsc{GraSP} can consistently achieve superior GED/MCS estimation performance over all the baselines while retaining high efficiency.

In summary, our contributions are as follows:

$\bullet$ We propose a new approach \textsc{GraSP} for GSC, which offers remarkable performance by virtue of several designs, including techniques for feature enhancement via positional encoding and multi-scale pooling over node representations.

$\bullet$ We theoretically analyze and reveal that \textsc{GraSP} can generate expressive representations that are beyond the 1-WL graph isomorphism test while achieving efficient time complexity with respect to the size of a graph pair.

$\bullet$ The superior performance of \textsc{GraSP}, in terms of efficiency and effectiveness, is validated against various competitors over multiple benchmark datasets.

\section{Preliminaries} \label{section: 2}

\begin{figure}[t!] \centering
\begin{minipage}[c]{0.6\columnwidth}
  \centering
  \includegraphics[height=1.16cm]{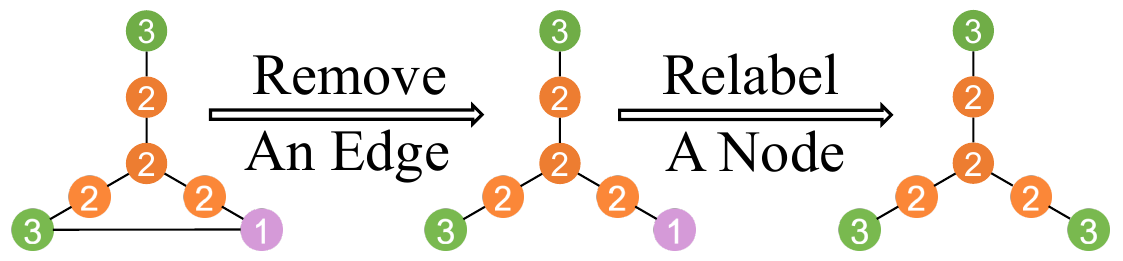}
\end{minipage}
\hfill
\begin{tikzpicture}
  \useasboundingbox (0,0);
  \draw[dashed] (0,0.7) -- (0,-0.7);
\end{tikzpicture}
\hfill
\begin{minipage}[c]{0.36\columnwidth}
  \centering
  \includegraphics[height=1.16cm]{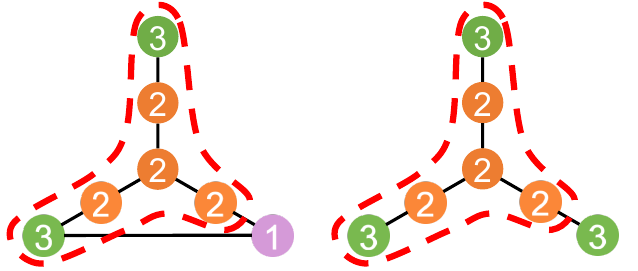}
\end{minipage}
\vspace{-1mm}
\caption{GED and MCS examples from AIDS700nef dataset. Left: GED is 2 and right: MCS is 6.}
\label{fig: 2.1}
\vspace{-3mm}
\end{figure}
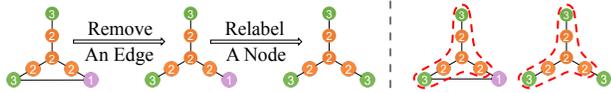

A graph $\mathcal{G}$ consists of a set of nodes $\mathcal{V}$ and a set of edges $\mathcal{E}$, i.e., $\mathcal{G} = (\mathcal{V}, \mathcal{E})$. The number of nodes and edges are $|\mathcal{V}|$ and $|\mathcal{E}|$, respectively. %
Given a graph database $\mathcal{D}$ containing a collection of graphs, we aim to build an end-to-end model that accurately estimates the similarity values of graph pairs. The model is designed to predict multiple similarity/distance metrics, including Graph Edit Distance (GED) and Maximum Common Subgraph (MCS).
\begin{definition}[Graph Edit Distance \cite{bunke_inexact_1983}]
GED is the edit distance between two graphs $\mathcal{G}_1$ and $\mathcal{G}_2$, i.e., the minimum number of edit operations to convert $\mathcal{G}_1$ to $\mathcal{G}_2$, where edit operations include adding/removing a node, adding/removing an edge, and relabeling a node. 
\end{definition}
\begin{definition}[Maximum Common Subgraph \cite{bunke_graph_1998}]
MCS is an isomorphic graph to the largest subgraph shared by two graphs $\mathcal{G}_1$ and $\mathcal{G}_2$.
\end{definition}
\noindent Following the choice of \cite{bai_learning-based_2020}, the MCS is connected. Figure \ref{fig: 2.1} shows GED and MCS examples.
Let $m$ be the number of labels, and we can define the node feature matrix $\mathbf{X} \in \mathbb{R}^{|\mathcal{V}| \times m}$ of a graph $\mathcal{G}$. 

During the training phase, 
a training sample  
consisting of a set of graph pair $(\mathcal{G}_1, \mathcal{G}_2) \in \mathcal{D} \times \mathcal{D} $ with ground-truth similarity value  $s(\mathcal{G}_1, \mathcal{G}_2)$. All training samples are used to train a prediction model. In the inference phase, the model predicts the similarity/distance of unseen graph pairs.

\section{Our Approach: \textsc{GraSP}} \label{section: 3}  
\begin{figure*}[!t] \centering
  \includegraphics[width=1.00\textwidth]{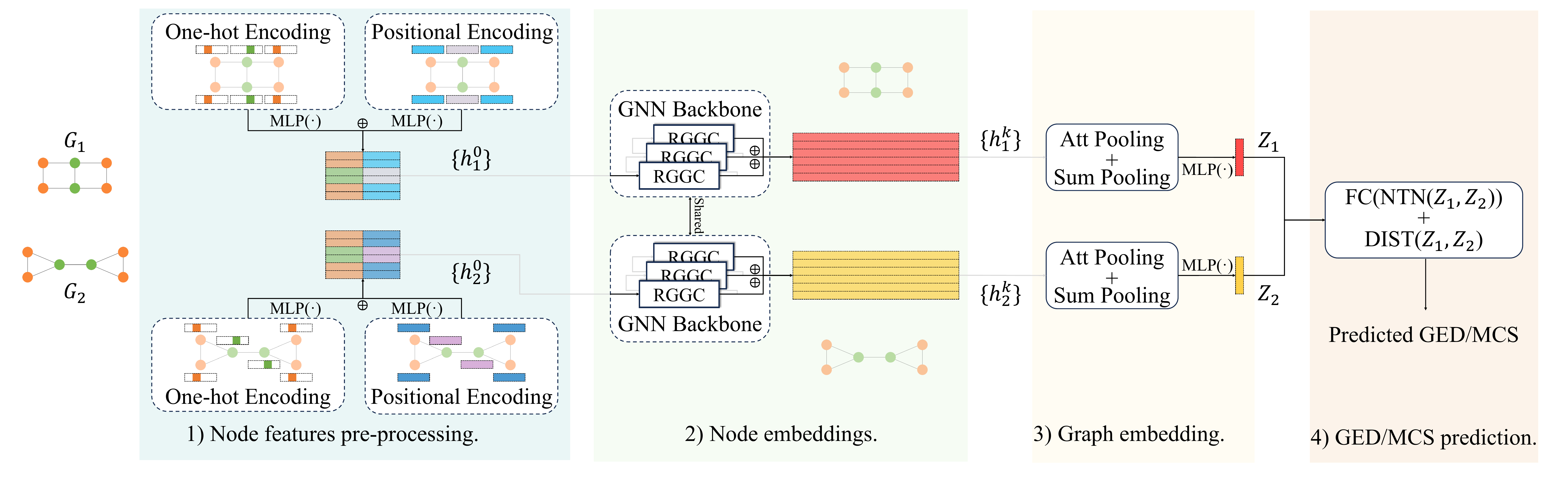}
  \vspace{-3mm}
  \caption{The architecture of \textsc{GraSP}.}
  \label{fig: 3.1}
  \vspace{-3mm}
\end{figure*}

The architecture of \textsc{GraSP} is illustrated in 
Figure \ref{fig: 3.1}.
In the node feature pre-processing, we first enhance node features by concatenating positional encoding of the node. This enhancement considers global graph topological features. Second, the node embedding module preserves the enhanced node features via an RGGC backbone into node embeddings. Third, within the graph embedding module, we devise a multi-scale node embedding pooling technique to obtain graph embeddings. Lastly, the prediction module predicts the similarity/distance between two graph embeddings.

\subsection{Enhanced Node Features via Positional Encoding} \label{section: 3.1}
Previous works like \cite{bai_simgnn_2020, bai_learning-based_2020, ranjan_greed_2023} obtain features using the one-hot encoding of node labels. Specifically, every node $i$ has $\mathbf{x}_i\in \mathbb{R} ^ {m}$ to represent its label, where $m$ is the number of labels in the graph database. 
Then $\mathbf{x}_i$ is transformed by an MLP to $\boldsymbol{\mu}_i \in \mathbb{R} ^ {d}$ \cite{ranjan_greed_2023}, which is then used as the input of GNN layers. 
However, only using node labels as node features to be the input of {message-passing graph neural networks} (MPGNNs) cannot pass the 1-WL graph isomorphism test. 
Specifically, MPGNNs update node representations based on message passing and by aggregating neighbor information, a node $i$ in a graph can be represented as:
\begin{equation}
    \mathbf{h}_i^{(\ell)} = f\left(\mathbf{h}_i^{(\ell - 1)}, \left\{\mathbf{h}_j^{(\ell-1)} \middle| j \in \mathcal{N}(i)\right\}\right),
    \label{eq. 1}
\end{equation}
where $\ell$ represents $\ell$-th layer of the stack of the MPGNN layers, $\mathcal{N}(i)$ denotes the neighbors of node $i$. 

\begin{figure}[!t] \centering
\includegraphics[width=0.5\columnwidth]{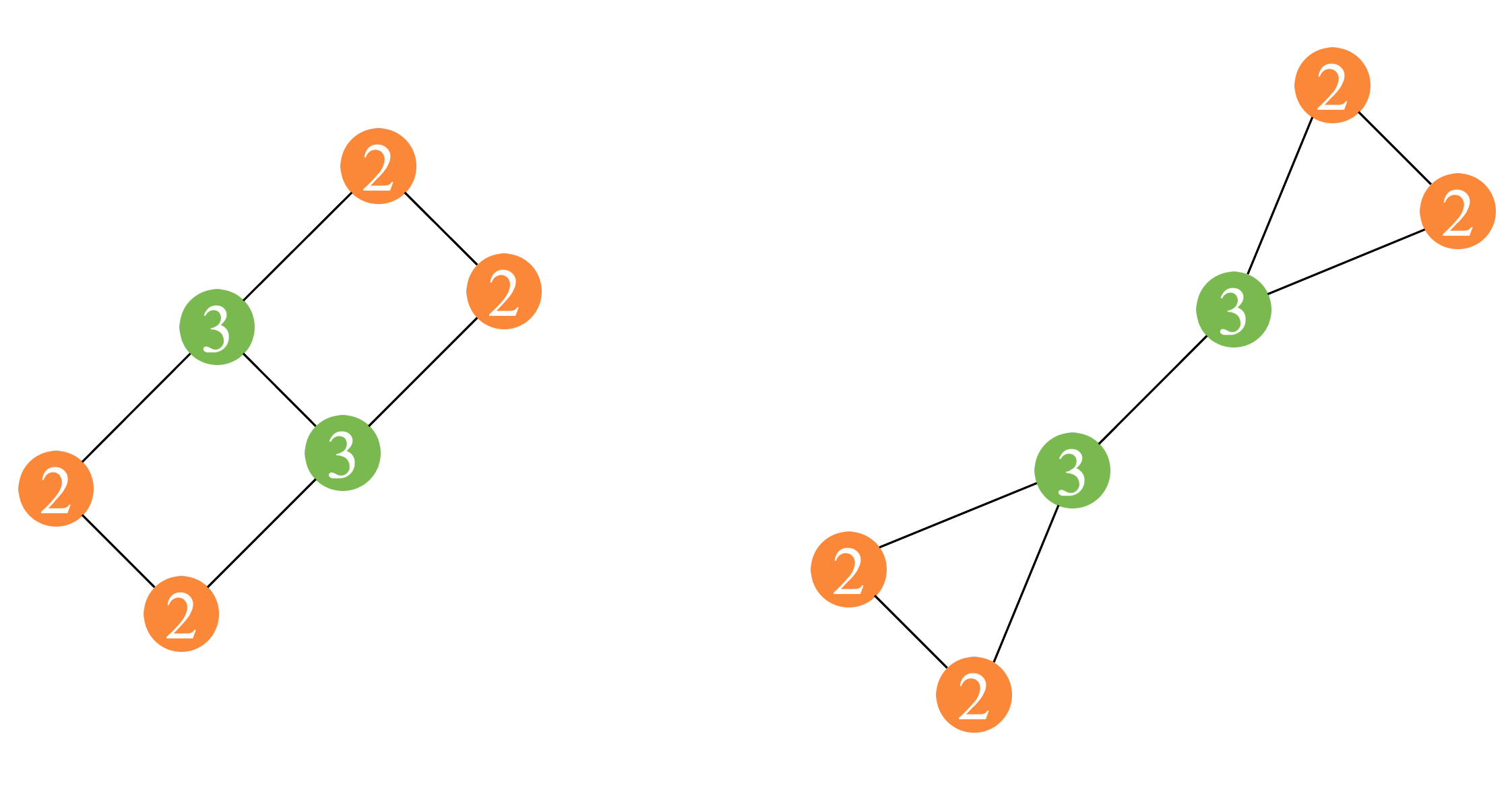}
\vspace{-3mm}
    \caption{An example that MPGNNs fail to distinguish.}
    \label{fig: 3.2}
    \vspace{-3mm}
\end{figure}
It has been shown by \cite{xu_how_2019} that MPGNNs can perform up to, but not beyond, the level of the 1-WL test. In general, the 1-WL test can effectively distinguish non-isomorphic graphs. However, since MPGNNs only aggregate information from their direct neighbors, they fail to capture higher-order structural properties and the global topology of the graph, thus limiting the expressive capability of MPGNNs. We take one case that MPGNNs fail to distinguish \cite{sato_survey_2020}, as shown in Figure \ref{fig: 3.2}. MPGNNs produce the same set of node embeddings for these two non-isomorphic graphs. Thus, the same set of node embeddings causes a model to give incorrect predictions that the GED is 0, which apparently should not be 0.

To mitigate this issue and enhance the expressiveness of our method, we propose to enhance node features with positional encoding that can exceed the expressive capacity of the 1-WL test. 
Specifically, we use the random walk method for position encoding, RWPE, which has been 
empirically proven to work by \cite{dwivedi_graph_2022}. Unlike the distance encoding method proposed by \cite{li_distance_2020} that uses random walks to learn the distance of all pairs of nodes in the graph, RWPE uses only the probability of a node landing on itself, i.e., the diagonal part of the random walk matrix. Given the random walk length $k$, we can use RWPE to precompute the positional features $\mathbf{p}_{i_{\textrm{init}}} \in \mathbb{R} ^ {k}$ of a node $i$ in the graph, denoted as $ \mathbf{p}_{i_{\textrm{init}}} = \left\{\mathbf{RW}_{ii}^{\ell} \middle| \ell=1,2,... ,k\right\}$, where $\mathbf{RW} = \mathbf{A}\mathbf{D}^{-1}$ is the random walk matrix obtained by adjacency matrix $\mathbf{A}$ and diagonal degree matrix $\mathbf{D}$ of a graph. 
Then we transform the RWPE $\mathbf{p}_{i_{\textrm{init}}}$ into $\mathbf{p}_i \in \mathbb{R} ^ {d}$ by an MLP.

We concatenate the transformed positional encoding $\mathbf{p}_i$ and the encoding $\boldsymbol{\mu}_i$  of node $i$ to  get its enhanced representation  $\mathbf{h}_{i}^{(0)} \in \mathbb{R} ^ {2d}$, to be the input of GNN backbone,
\begin{equation}
    \mathbf{h}_{i}^{(0)} = \textrm{CONCAT}\left(\boldsymbol{\mu}_i, \mathbf{p}_i\right).
    \label{eq. 2}
\end{equation}

\subsection{Multi-Scale Pooling on RGGC} \label{section: 3.2}
The enhanced node features $\mathbf{h}_{i}^{(0)}$ obtained in previous subsection
are then fed into a graph neural network consisting of $n$ layers of ResGatedGraphConv (RGGC layers) \cite{bresson_residual_2018} to learn the hidden representations of the nodes. %
The RGGC layers can leverage a gating mechanism where gating units learn to control information flow through the network. Moreover, residual connections are used in RGGC layers to help with gradient flow during training. With these two techniques incorporated, the RGGC layers can learn complex patterns in graph data, and therefore, are more powerful and versatile than basic graph convolutional layers like GCNs and GINs. 
At the $\ell$-th layer, the node representation $\mathbf{h}_{i}^{(\ell)}$ is
\begin{multline}\label{eq. 3}
    \mathbf{h}_{i}^{(\ell)} = \mathbf{h}_{i}^{(\ell - 1)} \\
    + \textrm{ReLU}(\mathbf{W}_{S}\mathbf{h}_{i}^{(\ell - 1)} + \sum _{j \in \mathcal{N}(i)} \eta_{i,j} \odot \mathbf{W}_{N}\mathbf{h}_{j}^{(\ell - 1)}),
\end{multline}
where $\mathbf{W}_{S} \in \mathbb{R} ^ {2d \times 2d}$ and $\mathbf{W}_{N} \in \mathbb{R} ^ {2d \times 2d}$ are learnable weight matrices, $\odot$ is the Hadamard point-wise multiplication operator and the gate $\eta_{i,j}$ is defined as $ \eta_{i,j} = \sigma(\mathbf{W}_{GS}\mathbf{h}_{i}^{(\ell - 1)} + \mathbf{W}_{GN}\mathbf{h}_{j}^{(\ell - 1)})$, where $\mathbf{W}_{GS} \in \mathbb{R} ^ {2d \times 2d}$, $\mathbf{W}_{GN} \in \mathbb{R} ^ {2d \times 2d}$ are learnable weight matrices and with $\sigma$ as an activation function. In \cite{bresson_residual_2018}, the sigmoid function is chosen as the activation function so that the gate $\eta_{i,j}$ can learn the weight controlling how important the information from node $j$ to node $i$. 

We concatenate the node representations obtained in all layers to preserve the information of multi-hop neighbors better. The concatenated representation of node $i$ after $n$ layers is $\mathbf{h}_{i} \in \mathbb{R} ^ {2(n+1)d}$, 
\begin{equation}
    \mathbf{h}_{i} = \textrm{CONCAT}\left(\left\{\mathbf{h}_{i}^{(\ell)} \middle| \ell=0,1,...,n\right\}\right).
    \label{eq. 4}
\end{equation}

Note that our task is to estimate the similarity between two graphs. Therefore, we need to generate graph-level representations based on the node representations above.
We design a {multi-scale pooling} technique that considers both attention pooling and summation pooling. 
Attention pooling \cite{bai_simgnn_2020} assigns weights to each node according to its importance, and then pools the node embeddings using a weighted sum based on the attention weights. Denote the resulting graph embedding as $\mathbf{z}_\textrm{att} \in \mathbb{R} ^ {2nd}$. We get 
$$\mathbf{z}_{\textrm{att}} = \sum_{i=1}^{|\mathcal{V}|} \sigma(\mathbf{h}_i^T\tanh(\frac{1}{{|\mathcal{V}|}}\mathbf{W}_{A}\sum_{j=1}^{|\mathcal{V}|} \mathbf{h}_j))\mathbf{h}_i, $$where $\mathbf{W}_{A} \in \mathbb{R} ^ {\left[2(n+1)d\right] \times \left[2(n+1)d\right]}$ is learnable and $\sigma$ is a sigmoid function. Attention pooling employs an attention mechanism and therefore one merit is that it is globally context-aware. 
Summation pooling $\mathbf{z}_\textrm{sum} \in \mathbb{R} ^ {2(n+1)d}$  simply sums the node embeddings, i.e., $\mathbf{z}_{\textrm{sum}} = \sum_{i=1}^{|\mathcal{V}|} \mathbf{h}_i$. Summation pooling has the advantage of simplicity and thus is more computationally efficient.

We observe that both of the above pooling methods have some drawbacks: summation pooling treats all nodes equally, which may not be optimal; and attention in attention pooling runs the risk of overfitting on specific nodes. 
Therefore, in our multi-scale pooling, we 
mix these two pooling methods and let the model learn to trade off the two pooling operations, thus reducing the drawbacks of the two pooling methods. 
We get the combined graph embedding $ \mathbf{z} \in \mathbb{R} ^ {2(n+1)d}$:
\begin{equation}
\mathbf{z}_{\textrm{combined}} = \mathbf{a} \mathbf{z}_{\textrm{att}} + \left(1 - \mathbf{a}\right)\mathbf{z}_{\textrm{sum}},
\end{equation}
where $\mathbf{a} \in \mathbb{R} ^{2(n+1)d}$  is a vector that can be learned. 

Similar to the node feature pre-processing, we pass the graph embedding that has gone through the pooling layer via an MLP to adjust its dimension for subsequent processing, and finally, we get the graph embedding $\mathbf{z} \in \mathbb{R} ^ {d}$.

\subsection{GED and MCS Prediction Objectives} \label{section: 3.3}
After obtaining the graph embeddings $\mathbf{z}_1$ and $\mathbf{z}_2$ of two graphs $\mathcal{G}_1$ and $\mathcal{G}_2$, we explain how to obtain predicted GED values and the design of training objective. The way to get MCS estimation and training objectives naturally follows. 

To get a predicted GED, an intuitive idea is to compute the Euclidean distance between $\mathbf{z}_1$ and $\mathbf{z}_2$, 
\begin{equation} \label{eq. 6}
    \textrm{dist}(\mathbf{z}_1,\mathbf{z}_2)=\|\mathbf{z}_1 - \mathbf{z}_2\|_2.
\end{equation}

Moreover, inspired by \cite{zhuo_efficient_2022}, we introduce the Neural Tensor Network (NTN) \cite{socher_reasoning_2013} as a multi-headed weighted cosine similarity function to compute the interaction value of the two graph embeddings in the capacity of a bias value as a complement to Eq.~\eqref{eq. 6}. NTN is a powerful method for quantifying relationships between representations. We denote the interaction value of embeddings $\mathbf{z}_1$ and $\mathbf{z}_2$ as:
\begin{multline}
    \textrm{interact}(\mathbf{z}_1, \mathbf{z}_2) = \textrm{MLP}(\textrm{ReLU}(\mathbf{z}_1^T\mathbf{W}_{I}^{\left[1:t\right]}\mathbf{z}_2 \\
    +\mathbf{W}_{C}\textrm{CONCAT}(\mathbf{z}_1,\mathbf{z}_2)+\mathbf{b})),
    \label{eq. 7}
\end{multline}
\noindent where $\mathbf{W}_{I}^{\left[1:t\right]} \in \mathbb{R} ^ {d \times d \times t}$ is a learnable weight tensor, $\mathbf{W}_{C}$ is a learnable weight matrix, $\mathbf{b} \in \mathbb{R} ^ t$ is a bias vector, $t$ is the hyperparameter controlling the NTN output and MLP(·) is a fully connected neural network that maps the similarity from $\mathbb{R} ^ {t}$ to $\mathbb{R}$. 

Finally, our predicted GED value is
\begin{equation} \label{eq. 8}
\small
    \textrm{GED}(\mathcal{G}_1, \mathcal{G}_2) = \boldsymbol{\beta}\cdot \textrm{dist}(\mathbf{z}_1,\mathbf{z}_2)
    +(1-\boldsymbol{\beta})\cdot  \textrm{interact}(\mathbf{z}_1, \mathbf{z}_2),
\end{equation}
where $\boldsymbol{\beta}$ is a scalar that can be learned. 

Then we adopt  the mean squared error between our predicted GED value $\textrm{GED}(\mathcal{G}_1, \mathcal{G}_2)$ and ground-truth GED value $\textrm{GED}^*(\mathcal{G}_1, \mathcal{G}_2)$, and have the loss function:
\begin{equation} \label{eq. 9}
\small
    \mathcal{L}= \frac{1}{T}\sum_{(\mathcal{G}_1,\mathcal{G}_2) \in \mathcal{D} \times \mathcal{D}}\textrm{MSE}\left(\textrm{GED}(\mathcal{G}_1, \mathcal{G}_2),\textrm{GED}^*(\mathcal{G}_1, \mathcal{G}_2)\right),
\end{equation}
\noindent where $T$ is the number of training graph pairs in a graph database $\mathcal{D}$ and value  $\textrm{GED}^*(\mathcal{G}_1, \mathcal{G}_2)$ is the ground-truth GED  between graph $\mathcal{G}_1$ and $\mathcal{G}_2$.

Eq.~\eqref{eq. 8} can be generalized to other similarity metrics like MCS. Recall that MCS is a similarity measuring the largest common subgraph of two graphs. Therefore, we can consider the output of interaction, i.e., Eq.~\eqref{eq. 7} as the similarity of the two graph embeddings and further consider Eq.~\eqref{eq. 6} as a bias value. Therefore, we keep the right-hand side of Eq.~\eqref{eq. 8} unchanged and change the left-hand side to $\textrm{MCS}(\mathcal{G}_1, \mathcal{G}_2)$. The loss function in Eq.~\eqref{eq. 9} follows for MCS.

\section{Analysis of \textsc{GraSP}} \label{section: 4}

We first prove that \textsc{GraSP} achieves high expressiveness and can pass the 1-WL test, and then analyze the complexity of \textsc{GraSP} that is linear to the number of nodes in a graph pair.

We use the following Proposition \ref{prop: power} to formally state that our method can outperform the 1-WL test. The proof is provided in 
the Appendix.
We utilize the proof in \cite{xu_how_2019} that the representation ability of the 1-WL test is equivalent to that of standard MPGNNs in the graph isomorphism problem.

\begin{proposition}\label{prop: power}
Given a pair of non-isomorphic graphs $\mathcal{G}_1$ and $\mathcal{G}_2$ that cannot be discriminated by the 1-WL test, and the two graphs have different sets of initial position encodings, then  \textsc{GraSP} can generate different graph representations for the two non-isomorphic graphs.
\end{proposition}

 The preconditions of Proposition \ref{prop: power} include that two graphs should have different initial position encodings. According to the definition of RWPE, nodes on non-isomorphic graphs generally get different sets of RWPEs when $k$ is sufficiently large \cite{dwivedi_graph_2022}. Thus, RWPE satisfies this precondition that the sets of positional encodings are different.
Therefore, graph embeddings that are more discriminative than the 1-WL test are obtained. These powerful graph embeddings can benefit our graph similarity prediction task when predicting GED and MCS objectives.

\noindent\textbf{Complexity Analysis.} The inference time complexity of \textsc{GraSP} is linear to the number of nodes of the graph pair. Our node feature preprocessing module requires downscaling the dimensionality of the node features from $\mathbb{R} ^ {m}$ to $\mathbb{R} ^ {d}$, resulting in a time complexity of $O(md|\mathcal{V}|)$. The positional encoding module contains a random walk positional encoding pre-computation and an MLP. 
The pre-computation of random walk positional encoding takes O($k|\mathcal{V}|^2$) due to the sparse matrix multiplication. This pre-computation is performed only once when the number of steps $k$ for the random walker is determined. Since it can be reused across both training and inference stages, it is regarded as a preprocessing technique and excluded from the complexity analysis. The MLP(·) takes $O(kd|\mathcal{\mathcal{V}}|)$ time. 
The node embedding module contains $n$ layers of RGGC with a time complexity of $O(n|\mathcal{E}|)$. The multi-scale pooling module contains attention pooling and summation pooling, both with time complexity of $O(nd|\mathcal{V}|)$. 
In the phase of generating the final graph embedding, we downscale the dimensionality of the graph embedding from $\mathbb{R} ^ {2nd}$ to $\mathbb{R} ^ {d}$ with a time complexity of $O(nd^2)$. In the similarity prediction module, the time complexity of NTN interaction is $O(d^2t)$, where $t$ is the dimension of NTN output, and the time complexity of Euclidean distance calculation is $O(d)$  and the total time is $O(d^2t)$.
Hence, the time complexity of \textsc{GraSP} for predictions is $O(d|\mathcal{V}|(m+ k + n) + nd^2+d^2t)$, which is linear to the number of nodes of the graph pair.

\begin{table*}[ht]
\centering
\setlength\tabcolsep{3pt} 
\resizebox{1.86\columnwidth}{!}{
\begin{tabular}{@{}lcccccccccc@{}}
\toprule
\multicolumn{1}{l}{} & \multicolumn{5}{c}{AIDS700nef}                                       & \multicolumn{5}{c}{IMDBMulti}                                           \\ \cmidrule(lr){2-6} \cmidrule(lr){7-11} 
\multicolumn{1}{l}{} & MAE $\downarrow$      & $\rho \uparrow$         & $\tau \uparrow$          & P@10 $\uparrow$       & P@20 $\uparrow$       & MAE $\downarrow$      & $\rho \uparrow$         & $\tau \uparrow$          & P@10 $\uparrow$       & P@20 $\uparrow$         \\ \midrule
\textsc{SimGNN}               & 0.852\scriptsize$\pm$0.033 & 0.838\scriptsize$\pm$0.005 & 0.659\scriptsize$\pm$0.006 & 0.489\scriptsize$\pm$0.027 & 0.600\scriptsize$\pm$0.018 & 5.836\scriptsize$\pm$0.439 & 0.924\scriptsize$\pm$0.003 & 0.811\scriptsize$\pm$0.006 & 0.802\scriptsize$\pm$0.033 & 0.819\scriptsize$\pm$0.021 \\
\textsc{GraphSim}             & 0.979\scriptsize$\pm$0.019 & 0.808\scriptsize$\pm$0.012 & 0.654\scriptsize$\pm$0.008 & 0.382\scriptsize$\pm$0.021 & 0.502\scriptsize$\pm$0.013 & 7.907\scriptsize$\pm$0.670 & 0.738\scriptsize$\pm$0.012 & 0.621\scriptsize$\pm$0.009 & 0.578\scriptsize$\pm$0.028 & 0.588\scriptsize$\pm$0.023 \\
GMN                  & 0.881\scriptsize$\pm$0.010 & 0.831\scriptsize$\pm$0.003 & 0.650\scriptsize$\pm$0.004 & 0.461\scriptsize$\pm$0.010 & 0.573\scriptsize$\pm$0.006 & 5.101\scriptsize$\pm$0.860 & 0.925\scriptsize$\pm$0.006 & 0.814\scriptsize$\pm$0.008 & 0.830\scriptsize$\pm$0.018 & 0.855\scriptsize$\pm$0.010 \\
MGMN                 & 0.805\scriptsize$\pm$0.006 & 0.863\scriptsize$\pm$0.001 & 0.727\scriptsize$\pm$0.002 & 0.553\scriptsize$\pm$0.010 & 0.637\scriptsize$\pm$0.004 & 12.876\scriptsize$\pm$0.728 & 0.771\scriptsize$\pm$0.029 & 0.669\scriptsize$\pm$0.023 & 0.307\scriptsize$\pm$0.033 & 0.353\scriptsize$\pm$0.024 \\
H2MN                 & 0.731\scriptsize$\pm$0.014 & 0.875\scriptsize$\pm$0.002 & 0.740\scriptsize$\pm$0.003 & 0.552\scriptsize$\pm$0.008 & 0.652\scriptsize$\pm$0.010 & 9.735\scriptsize$\pm$1.542 & 0.909\scriptsize$\pm$0.010 & 0.809\scriptsize$\pm$0.012 & 0.746\scriptsize$\pm$0.057 & 0.783\scriptsize$\pm$0.032 \\
EGSC                 & \underline{0.667\scriptsize$\pm$0.010} & \underline{0.889\scriptsize$\pm$0.001} & 0.720\scriptsize$\pm$0.002 & \underline{0.662\scriptsize$\pm$0.008} & \underline{0.735\scriptsize$\pm$0.004} & \underline{4.397\scriptsize$\pm$0.101} & \underline{0.940\scriptsize$\pm$0.001} & 0.847\scriptsize$\pm$0.002 & \underline{0.861\scriptsize$\pm$0.009} & \underline{0.874\scriptsize$\pm$0.007} \\
ERIC                 & 0.765\scriptsize$\pm$0.016 & 0.876\scriptsize$\pm$0.003 & 0.746\scriptsize$\pm$0.005 & 0.592\scriptsize$\pm$0.015 & 0.681\scriptsize$\pm$0.015 & 5.487\scriptsize$\pm$0.929 & 0.934\scriptsize$\pm$0.003 & \underline{0.862\scriptsize$\pm$0.006} & 0.837\scriptsize$\pm$0.014 & 0.836\scriptsize$\pm$0.012 \\
\textsc{Greed}                & 0.731\scriptsize$\pm$0.013 & 0.886\scriptsize$\pm$0.005 & \underline{0.758\scriptsize$\pm$0.006} & 0.617\scriptsize$\pm$0.013 & 0.709\scriptsize$\pm$0.015 & 4.418\scriptsize$\pm$0.285 & 0.930\scriptsize$\pm$0.003 & 0.857\scriptsize$\pm$0.004 & 0.837\scriptsize$\pm$0.010 & 0.850\scriptsize$\pm$0.003 \\
NA-GSL               & 0.765\scriptsize$\pm$0.014 & 0.881\scriptsize$\pm$0.001 & 0.753\scriptsize$\pm$0.002 & 0.562\scriptsize$\pm$0.023 & 0.665\scriptsize$\pm$0.010 & 7.588\scriptsize$\pm$0.975 & 0.900\scriptsize$\pm$0.009 & 0.814\scriptsize$\pm$0.009 & 0.544\scriptsize$\pm$0.049 & 0.611\scriptsize$\pm$0.033 \\
GEDGNN               & 0.724\scriptsize$\pm$0.012 & 0.885\scriptsize$\pm$0.002 & 0.757\scriptsize$\pm$0.003 & 0.604\scriptsize$\pm$0.011 & 0.685\scriptsize$\pm$0.012 & 8.585\scriptsize$\pm$1.494 & 0.902\scriptsize$\pm$0.006 & 0.834\scriptsize$\pm$0.014 & 0.748\scriptsize$\pm$0.008 & 0.771\scriptsize$\pm$0.026 \\
\textsc{GraSP}       & \textbf{0.640\scriptsize$\pm$0.013} & \textbf{0.917\scriptsize$\pm$0.002} & \textbf{0.804\scriptsize$\pm$0.003} & \textbf{0.741\scriptsize$\pm$0.009} & \textbf{0.800\scriptsize$\pm$0.002} & \textbf{3.966\scriptsize$\pm$0.064} & \textbf{0.942\scriptsize$\pm$0.001} & \textbf{0.874\scriptsize$\pm$0.003} & \textbf{0.863\scriptsize$\pm$0.011} & \textbf{0.876\scriptsize$\pm$0.014} \\
\midrule \midrule
\multicolumn{1}{l}{} & \multicolumn{5}{c}{LINUX}                                            & \multicolumn{5}{c}{PTC}                                                 \\ \cmidrule(lr){2-6} \cmidrule(lr){7-11} 
\multicolumn{1}{l}{} & MAE $\downarrow$      & $\rho \uparrow$         & $\tau \uparrow$          & P@10 $\uparrow$       & P@20 $\uparrow$       & MAE $\downarrow$      & $\rho \uparrow$         & $\tau \uparrow$          & P@10 $\uparrow$       & P@20 $\uparrow$         \\ \midrule
\textsc{SimGNN}               & 0.269\scriptsize$\pm$0.027 & 0.939\scriptsize$\pm$0.002 & 0.792\scriptsize$\pm$0.004 & 0.947\scriptsize$\pm$0.012 & 0.959\scriptsize$\pm$0.016 & 4.200\scriptsize$\pm$0.395 & 0.930\scriptsize$\pm$0.006 & 0.821\scriptsize$\pm$0.008 & 0.465\scriptsize$\pm$0.028 & 0.628\scriptsize$\pm$0.031 \\
\textsc{GraphSim}             & 0.212\scriptsize$\pm$0.033 & 0.962\scriptsize$\pm$0.003 & 0.889\scriptsize$\pm$0.003 & 0.970\scriptsize$\pm$0.010 & 0.982\scriptsize$\pm$0.007 & 5.003\scriptsize$\pm$0.491 & 0.850\scriptsize$\pm$0.009 & 0.721\scriptsize$\pm$0.011 & 0.408\scriptsize$\pm$0.008 & 0.521\scriptsize$\pm$0.011 \\
GMN                  & 0.293\scriptsize$\pm$0.047 & 0.937\scriptsize$\pm$0.001 & 0.788\scriptsize$\pm$0.002 & 0.906\scriptsize$\pm$0.022 & 0.932\scriptsize$\pm$0.012 & 4.353\scriptsize$\pm$1.051 & \underline{0.941\scriptsize$\pm$0.005} & 0.836\scriptsize$\pm$0.008 & 0.558\scriptsize$\pm$0.041 & 0.675\scriptsize$\pm$0.028 \\
MGMN                 & 0.368\scriptsize$\pm$0.035 & 0.958\scriptsize$\pm$0.003 & 0.883\scriptsize$\pm$0.005 & 0.907\scriptsize$\pm$0.042 & 0.931\scriptsize$\pm$0.017 & 3.839\scriptsize$\pm$0.374 & 0.920\scriptsize$\pm$0.004 & 0.802\scriptsize$\pm$0.007 & 0.503\scriptsize$\pm$0.027 & 0.629\scriptsize$\pm$0.017 \\
H2MN                 & 0.909\scriptsize$\pm$0.523 & 0.961\scriptsize$\pm$0.004 & 0.874\scriptsize$\pm$0.006 & 0.951\scriptsize$\pm$0.016 & 0.953\scriptsize$\pm$0.016 & 5.391\scriptsize$\pm$2.151 & 0.918\scriptsize$\pm$0.012 & 0.809\scriptsize$\pm$0.017 & 0.435\scriptsize$\pm$0.047 & 0.562\scriptsize$\pm$0.035 \\
EGSC                 & 0.083\scriptsize$\pm$0.019 & 0.948\scriptsize$\pm$0.003 & 0.811\scriptsize$\pm$0.003 & 0.981\scriptsize$\pm$0.009 & 0.988\scriptsize$\pm$0.009 & \underline{3.743\scriptsize$\pm$0.203} & 0.938\scriptsize$\pm$0.002 & 0.835\scriptsize$\pm$0.003 & \underline{0.577\scriptsize$\pm$0.012} & \underline{0.695\scriptsize$\pm$0.008} \\
ERIC                 & \underline{0.080\scriptsize$\pm$0.011} & 0.971\scriptsize$\pm$0.002 & 0.906\scriptsize$\pm$0.003 & 0.977\scriptsize$\pm$0.005 & 0.983\scriptsize$\pm$0.006 & 3.908\scriptsize$\pm$0.099 & 0.940\scriptsize$\pm$0.002 & \underline{0.842\scriptsize$\pm$0.002} & 0.555\scriptsize$\pm$0.010 & 0.670\scriptsize$\pm$0.011 \\
\textsc{Greed}                & 0.342\scriptsize$\pm$0.006 & 0.963\scriptsize$\pm$0.001 & 0.884\scriptsize$\pm$0.001 & 0.964\scriptsize$\pm$0.004 & 0.970\scriptsize$\pm$0.003 & 3.934\scriptsize$\pm$0.132 & 0.911\scriptsize$\pm$0.004 & 0.797\scriptsize$\pm$0.005 & 0.432\scriptsize$\pm$0.015 & 0.545\scriptsize$\pm$0.017 \\
NA-GSL               & 0.132\scriptsize$\pm$0.023 & \underline{0.973\scriptsize$\pm$0.001} & \underline{0.914\scriptsize$\pm$0.001} & \textbf{0.987\scriptsize$\pm$0.005} & \underline{0.989\scriptsize$\pm$0.003} & \multicolumn{5}{c}{OOM} \\
GEDGNN               & 0.108\scriptsize$\pm$0.015 & 0.971\scriptsize$\pm$0.003 & 0.906\scriptsize$\pm$0.003 & 0.966\scriptsize$\pm$0.011 & 0.975\scriptsize$\pm$0.008 & 4.164\scriptsize$\pm$0.274 & 0.935\scriptsize$\pm$0.002 & 0.839\scriptsize$\pm$0.002 & 0.543\scriptsize$\pm$0.025 & 0.659\scriptsize$\pm$0.008 \\
\textsc{GraSP}       & \textbf{0.042\scriptsize$\pm$0.003} & \textbf{0.975\scriptsize$\pm$0.001} & \textbf{0.922\scriptsize$\pm$0.001} & \underline{0.983\scriptsize$\pm$0.003} & \textbf{0.992\scriptsize$\pm$0.002} & \textbf{3.556\scriptsize$\pm$0.080} & \textbf{0.952\scriptsize$\pm$0.002} & \textbf{0.861\scriptsize$\pm$0.003} & \textbf{0.612\scriptsize$\pm$0.019} & \textbf{0.707\scriptsize$\pm$0.010} \\
\bottomrule
\end{tabular}
}
\vspace{-2mm}
\caption{\label{table: 5.2.1}Effectiveness results on  GED predictions with standard deviation. \textbf{Bold}: best, \underline{Underline}: runner-up.} 
\vspace{-3mm}
\end{table*}

\begin{table*}[ht]
\centering
\setlength\tabcolsep{3pt} 
\resizebox{1.86\columnwidth}{!}{
\begin{tabular}{@{}lcccccccccc@{}}
\toprule
\multicolumn{1}{l}{} & \multicolumn{5}{c}{AIDS700nef}                                       & \multicolumn{5}{c}{IMDBMulti}                                           \\ \cmidrule(lr){2-6} \cmidrule(lr){7-11} 
\multicolumn{1}{l}{} & MAE $\downarrow$      & $\rho \uparrow$         & $\tau \uparrow$          & P@10 $\uparrow$       & P@20 $\uparrow$       & MAE $\downarrow$      & $\rho \uparrow$         & $\tau \uparrow$          & P@10 $\uparrow$       & P@20 $\uparrow$         \\ \midrule
\textsc{SimGNN}               & 0.572\scriptsize$\pm$0.019 & 0.796\scriptsize$\pm$0.004 & 0.615\scriptsize$\pm$0.004 & 0.610\scriptsize$\pm$0.014 & 0.662\scriptsize$\pm$0.014 & 0.284\scriptsize$\pm$0.086 & 0.783\scriptsize$\pm$0.005 & 0.618\scriptsize$\pm$0.006 & 0.889\scriptsize$\pm$0.014 & 0.918\scriptsize$\pm$0.016 \\
\textsc{GraphSim}             & 0.442\scriptsize$\pm$0.014 & 0.802\scriptsize$\pm$0.009 & 0.679\scriptsize$\pm$0.006 & 0.529\scriptsize$\pm$0.019 & 0.596\scriptsize$\pm$0.018 & 0.786\scriptsize$\pm$0.077 & 0.764\scriptsize$\pm$0.005 & 0.671\scriptsize$\pm$0.008 & 0.787\scriptsize$\pm$0.019 & 0.791\scriptsize$\pm$0.023 \\
GMN                  & 0.747\scriptsize$\pm$0.023 & 0.701\scriptsize$\pm$0.002 & 0.552\scriptsize$\pm$0.001 & 0.421\scriptsize$\pm$0.013 & 0.489\scriptsize$\pm$0.005 & 0.274\scriptsize$\pm$0.038 & 0.782\scriptsize$\pm$0.003 & 0.617\scriptsize$\pm$0.003 & 0.910\scriptsize$\pm$0.012 & 0.917\scriptsize$\pm$0.011 \\
MGMN                 & 0.510\scriptsize$\pm$0.011 & 0.841\scriptsize$\pm$0.007 & 0.723\scriptsize$\pm$0.007 & 0.658\scriptsize$\pm$0.018 & 0.686\scriptsize$\pm$0.010 & 0.758\scriptsize$\pm$0.068 & 0.770\scriptsize$\pm$0.002 & 0.686\scriptsize$\pm$0.003 & 0.884\scriptsize$\pm$0.025 & 0.845\scriptsize$\pm$0.023 \\
H2MN                 & 0.537\scriptsize$\pm$0.013 & 0.838\scriptsize$\pm$0.006 & 0.713\scriptsize$\pm$0.008 & 0.578\scriptsize$\pm$0.014 & 0.639\scriptsize$\pm$0.015 & 0.299\scriptsize$\pm$0.027 & 0.872\scriptsize$\pm$0.001 & 0.772\scriptsize$\pm$0.005 & 0.903\scriptsize$\pm$0.015 & 0.912\scriptsize$\pm$0.009 \\
EGSC                 & \underline{0.335\scriptsize$\pm$0.006} & 0.874\scriptsize$\pm$0.006 & 0.706\scriptsize$\pm$0.004 & \underline{0.798\scriptsize$\pm$0.003} & \underline{0.826\scriptsize$\pm$0.007} & \underline{0.147\scriptsize$\pm$0.012} & 0.792\scriptsize$\pm$0.007 & 0.636\scriptsize$\pm$0.005 & \underline{0.945\scriptsize$\pm$0.005} & \underline{0.956\scriptsize$\pm$0.005} \\
ERIC                 & 0.382\scriptsize$\pm$0.006 & \underline{0.895\scriptsize$\pm$0.008} & \underline{0.788\scriptsize$\pm$0.008} & 0.747\scriptsize$\pm$0.004 & 0.789\scriptsize$\pm$0.025 & 0.183\scriptsize$\pm$0.017 & 0.879\scriptsize$\pm$0.002 & \underline{0.796\scriptsize$\pm$0.002} & 0.919\scriptsize$\pm$0.012 & 0.938\scriptsize$\pm$0.008 \\
\textsc{Greed}                & 0.458\scriptsize$\pm$0.026 & 0.890\scriptsize$\pm$0.003 & 0.781\scriptsize$\pm$0.004 & 0.747\scriptsize$\pm$0.015 & 0.790\scriptsize$\pm$0.008 & 0.251\scriptsize$\pm$0.042 & 0.876\scriptsize$\pm$0.002 & 0.791\scriptsize$\pm$0.003 & 0.897\scriptsize$\pm$0.014 & 0.920\scriptsize$\pm$0.011 \\
NA-GSL               & 0.513\scriptsize$\pm$0.010 & 0.850\scriptsize$\pm$0.002 & 0.731\scriptsize$\pm$0.002 & 0.638\scriptsize$\pm$0.012 & 0.698\scriptsize$\pm$0.009 & 0.251\scriptsize$\pm$0.039 & \underline{0.877\scriptsize$\pm$0.002} & 0.794\scriptsize$\pm$0.003 & 0.902\scriptsize$\pm$0.016 & 0.908\scriptsize$\pm$0.015 \\
GEDGNN               & 0.439\scriptsize$\pm$0.019 & 0.890\scriptsize$\pm$0.005 & 0.780\scriptsize$\pm$0.003 & 0.718\scriptsize$\pm$0.009 & 0.779\scriptsize$\pm$0.010 & 0.348\scriptsize$\pm$0.017 & 0.872\scriptsize$\pm$0.003 & 0.781\scriptsize$\pm$0.014 & 0.883\scriptsize$\pm$0.026 & 0.889\scriptsize$\pm$0.022 \\
\textsc{GraSP}       & \textbf{0.330\scriptsize$\pm$0.007} & \textbf{0.923\scriptsize$\pm$0.002} & \textbf{0.822\scriptsize$\pm$0.002} & \textbf{0.848\scriptsize$\pm$0.007} & \textbf{0.872\scriptsize$\pm$0.005} & \textbf{0.134\scriptsize$\pm$0.018} & \textbf{0.881\scriptsize$\pm$0.001} & \textbf{0.797\scriptsize$\pm$0.002} & \textbf{0.947\scriptsize$\pm$0.007} & \textbf{0.957\scriptsize$\pm$0.004} \\
\midrule \midrule
\multicolumn{1}{l}{} & \multicolumn{5}{c}{LINUX}                                            & \multicolumn{5}{c}{PTC}                                                 \\ \cmidrule(lr){2-6} \cmidrule(lr){7-11} 
\multicolumn{1}{l}{} & MAE $\downarrow$      & $\rho \uparrow$         & $\tau \uparrow$          & P@10 $\uparrow$       & P@20 $\uparrow$       & MAE $\downarrow$      & $\rho \uparrow$         & $\tau \uparrow$          & P@10 $\uparrow$       & P@20 $\uparrow$         \\ \midrule
\textsc{SimGNN}               & 0.138\scriptsize$\pm$0.016 & 0.690\scriptsize$\pm$0.010 & 0.518\scriptsize$\pm$0.007 & 0.888\scriptsize$\pm$0.064 & 0.933\scriptsize$\pm$0.043 & 1.391\scriptsize$\pm$0.060 & 0.793\scriptsize$\pm$0.010 & 0.628\scriptsize$\pm$0.010 & 0.492\scriptsize$\pm$0.015 & 0.593\scriptsize$\pm$0.015  \\
\textsc{GraphSim}             & 0.287\scriptsize$\pm$0.034 & 0.788\scriptsize$\pm$0.017 & 0.699\scriptsize$\pm$0.004 & 0.810\scriptsize$\pm$0.021 & 0.823\scriptsize$\pm$0.034 & 1.789\scriptsize$\pm$0.071 & 0.794\scriptsize$\pm$0.012 & 0.648\scriptsize$\pm$0.014 & 0.448\scriptsize$\pm$0.025 & 0.531\scriptsize$\pm$0.028 \\
GMN                  & 0.187\scriptsize$\pm$0.016 & 0.678\scriptsize$\pm$0.007 & 0.506\scriptsize$\pm$0.007 & 0.876\scriptsize$\pm$0.038 & 0.904\scriptsize$\pm$0.023 & 1.610\scriptsize$\pm$0.075 & 0.735\scriptsize$\pm$0.015 & 0.571\scriptsize$\pm$0.011 & 0.462\scriptsize$\pm$0.014 & 0.539\scriptsize$\pm$0.015 \\
MGMN                 & 0.176\scriptsize$\pm$0.027 & 0.806\scriptsize$\pm$0.024 & 0.710\scriptsize$\pm$0.027 & 0.832\scriptsize$\pm$0.104 & 0.888\scriptsize$\pm$0.054 & 1.441\scriptsize$\pm$0.152 & 0.829\scriptsize$\pm$0.020 & 0.695\scriptsize$\pm$0.024 & 0.530\scriptsize$\pm$0.037 & 0.610\scriptsize$\pm$0.027 \\
H2MN                 & 0.196\scriptsize$\pm$0.027 & 0.803\scriptsize$\pm$0.002 & 0.690\scriptsize$\pm$0.003 & 0.901\scriptsize$\pm$0.060 & 0.936\scriptsize$\pm$0.037 & 1.360\scriptsize$\pm$0.122 & 0.816\scriptsize$\pm$0.013 & 0.683\scriptsize$\pm$0.013 & 0.516\scriptsize$\pm$0.007 & 0.617\scriptsize$\pm$0.017 \\
EGSC                 & 0.049\scriptsize$\pm$0.007 & 0.700\scriptsize$\pm$0.008 & 0.528\scriptsize$\pm$0.005 & 0.988\scriptsize$\pm$0.009 & 0.991\scriptsize$\pm$0.006 & \underline{1.170\scriptsize$\pm$0.127} & \underline{0.846\scriptsize$\pm$0.012} & 0.690\scriptsize$\pm$0.012 & \underline{0.603\scriptsize$\pm$0.016} & \underline{0.694\scriptsize$\pm$0.014} \\
ERIC                 & \underline{0.037\scriptsize$\pm$0.009} & \underline{0.812\scriptsize$\pm$0.001} & \underline{0.715\scriptsize$\pm$0.001} & \underline{0.991\scriptsize$\pm$0.005} & \underline{0.995\scriptsize$\pm$0.002} & 1.406\scriptsize$\pm$0.119 & 0.817\scriptsize$\pm$0.007 & 0.694\scriptsize$\pm$0.007 & 0.570\scriptsize$\pm$0.018 & 0.647\scriptsize$\pm$0.021 \\
\textsc{Greed}                & 0.101\scriptsize$\pm$0.021 & 0.811\scriptsize$\pm$0.001 & 0.711\scriptsize$\pm$0.001 & 0.978\scriptsize$\pm$0.007 & 0.988\scriptsize$\pm$0.006 & 1.306\scriptsize$\pm$0.086 & 0.824\scriptsize$\pm$0.008 & 0.697\scriptsize$\pm$0.009 & 0.578\scriptsize$\pm$0.007 & 0.654\scriptsize$\pm$0.012 \\
NA-GSL               & 0.085\scriptsize$\pm$0.021 & \underline{0.812\scriptsize$\pm$0.001} & \underline{0.715\scriptsize$\pm$0.001} & 0.985\scriptsize$\pm$0.006 & 0.991\scriptsize$\pm$0.002 & \multicolumn{5}{c}{OOM} \\
GEDGNN               & 0.089\scriptsize$\pm$0.014 & 0.809\scriptsize$\pm$0.005 & 0.711\scriptsize$\pm$0.003 & 0.943\scriptsize$\pm$0.028 & 0.969\scriptsize$\pm$0.011 & 1.493\scriptsize$\pm$0.113 & 0.811\scriptsize$\pm$0.015 & \underline{0.700\scriptsize$\pm$0.017} & 0.535\scriptsize$\pm$0.012 & 0.630\scriptsize$\pm$0.016 \\
\textsc{GraSP}       & \textbf{0.037\scriptsize$\pm$0.005} & \textbf{0.813\scriptsize$\pm$0.001} & \textbf{0.716\scriptsize$\pm$0.001} & \textbf{0.995\scriptsize$\pm$0.006} & \textbf{0.996\scriptsize$\pm$0.002} & \textbf{1.162\scriptsize$\pm$0.028} & \textbf{0.887\scriptsize$\pm$0.002} & \textbf{0.762\scriptsize$\pm$0.003} & \textbf{0.665\scriptsize$\pm$0.005} & \textbf{0.729\scriptsize$\pm$0.006} \\
\bottomrule
\end{tabular}
}
\vspace{-2mm}
\caption{\label{table: 5.2.2}Effectiveness results on  MCS predictions with standard deviation. \textbf{Bold}: best, \underline{Underline}: runner-up.}
\vspace{-3mm}
\end{table*}

\section{Experiments} \label{section: 5}
We evaluate  \textsc{GraSP} against competitors on GED and MCS prediction tasks on well-adopted benchmarking datasets.

\subsection{Experiment Setup} \label{section: 5.1}
\noindent
\textbf{Data.}
We conduct experiments on four real-world datasets, including AIDS700nef, LINUX, IMDBMulti \cite{bai_simgnn_2020} and PTC \cite{bai_learning-based_2020}. 
The statistics and descriptions of the four datasets can be found in 
the Appendix.
We split training, validation, and testing data with a ratio of 6:2:2 for all datasets and all methods by following the setting in \cite{bai_simgnn_2020} and \cite{bai_learning-based_2020}.
For small datasets AIDS and LINUX, we use A* to calculate ground-truth GEDs. For IMDBMulti and PTC, we follow the way in  \cite{bai_simgnn_2020} and use the minimum of the results of three approximation methods Beam \cite{neuhaus_fast_2006}, Hungarian \cite{riesen_approximate_2009} and VJ \cite{fankhauser_speeding_2011} to be ground-truth GED. 
We use the MCSPLIT \cite{mccreesh_partitioning_2017} algorithm to calculate the ground-truth MCS.

\noindent\textbf{Baseline Methods.} We compare with  ERIC \cite{zhuo_efficient_2022}, \textsc{Greed} \cite{ranjan_greed_2023}, NA-GSL \cite{tan_exploring_2023},  GEDGNN \cite{piao_computing_2023}, H2MN \cite{zhang_h2mn_2021}, EGSC \cite{qin_slow_2021},  \textsc{SimGNN} \cite{bai_simgnn_2020}, GMN \cite{li_graph_2019}, \textsc{GraphSim} \cite{bai_learning-based_2020}, and MGMN \cite{ling_multilevel_2021}. We use their official code provided by the authors and because some methods are designed to predict graph similarity scores, i.e., exponentially normalized GEDs instead of original GED, we modify the regression head part of their models to align with experiment settings. Moreover, since some methods are only implemented for GED, we extend their official code for the MCS objective. 
GENN-A* \cite{wang_combinatorial_2021} 
cannot scale to large data, IMDBMulti and PTC, and thus it is omitted.

\noindent
\textbf{Hyperparameters.} In our method, we use a search range of \{w/o, 8, 16, 24, 32\} for the step size $k$ of the RWPE, \{4, 6, 8, 10, 12\} for the number of layers $\ell$ of the GNN backbone, and \{16, 32, 64, 128, 256\} for the dimensionality $d$ of the node hidden representations and also the final graph embedding. Our full hyperparameter settings on the four datasets can be found in 
the Appendix.
For competitors, we conduct experiments according to their hyperparameter settings reported by their works.

\noindent
\textbf{Evaluation Metrics.}
We compare the performance of all methods by Mean Absolute Error (MAE) between the predicted and ground-truth GED and MCS scores, Spearman's Rank Correlation Coefficient ($\rho$) \cite{spearman_rho_1987}, Kendall's Rank Correlation Coefficient ($\tau$) \cite{kendall_tau_1938}, and Precision at 10 and 20 (P@10 and 20). The lower MAE proves that the model performs better; the higher the latter four, the better the model performs. 
Note that all these metrics are evaluated over ground-truth GED and MCS scores, instead of normalized graph similarity scores.

All experiments are conducted on a Linux machine with a  CPU Intel(R) Xeon(R) Gold 6226R CPU\@2.90GHz, and GPU model NVIDIA GeForce RTX 3090.

\subsection{Effectiveness} \label{section: 5.2}

Tables \ref{table: 5.2.1} and \ref{table: 5.2.2} report the overall effectiveness of all methods on all datasets for GED and MCS predictions, respectively. 
OOM means out of GPU memory.
In Table \ref{table: 5.2.1}, for GED predictions, our method \textsc{GraSP} outperforms all methods by all metrics on AIDS700nef, IMDBMulti and PTC datasets, and by 4 out of 5 metrics on LINUX dataset.
For example, On AIDS700nef, our method \textsc{GraSP} achieves high P@10 0.741, significantly outperforming the best competitor performance 0.662 of EGSC by 11.9\% relative improvement.
On LINUX, our method \textsc{GraSP} significantly reduces MAE to 0.042, compared with the best competitor ERIC with 0.080 MAE. 
In Table \ref{table: 5.2.2} for MCS predictions, our method \textsc{GraSP} outperforms all methods by all metrics on all four datasets.
For example, On PTC, \textsc{GraSP} can achieve 0.762 for $\tau$ metric, which is much higher than the runner-up GEDGNN with 0.7.
Moreover, the P@10 of \textsc{GraSP} is 0.665 while the best competitor EGSC has 0.603. 

The overall results in Tables \ref{table: 5.2.1} and \ref{table: 5.2.2} demonstrate the superior power of \textsc{GraSP} with simple but effective designs, including positional encoding enhanced node features and multi-scale pooling on RGGC backbone. 
Further, for intuitive results, interested readers may refer to 
the Appendix
for the absolute GED error heatmaps produced by \textsc{GraSP}, and recent baselines.

\begin{table}[!t]
    \centering
    \resizebox{\columnwidth}{!}{
    \renewcommand\tabcolsep{3pt}
    \begin{tabular}{@{}lccccc@{}}
    \toprule
                                  & MAE                             & $\rho$                          & $\tau$                          & P@10                            & P@20                            \\ \midrule
    \textsc{GraSP} (GIN)                    & 0.677\scriptsize$\pm$0.013 & 0.908\scriptsize$\pm$0.004 & 0.790\scriptsize$\pm$0.007 & 0.713\scriptsize$\pm$0.020 & 0.774\scriptsize$\pm$0.011 \\
    \textsc{GraSP} (GCN)                    & 0.654\scriptsize$\pm$0.011 & 0.911\scriptsize$\pm$0.003 & 0.795\scriptsize$\pm$0.004 & 0.721\scriptsize$\pm$0.006 & 0.783\scriptsize$\pm$0.009 \\
    \textsc{GraSP} (w/o pe)                 & \underline{0.651\scriptsize$\pm$0.016} & \underline{0.915\scriptsize$\pm$0.002} & \underline{0.800\scriptsize$\pm$0.002} & 0.726\scriptsize$\pm$0.010 & 0.793\scriptsize$\pm$0.006 \\
    \textsc{GraSP} (w/o att)                & 0.662\scriptsize$\pm$0.002 & 0.912\scriptsize$\pm$0.003 & 0.797\scriptsize$\pm$0.004 & 0.732\scriptsize$\pm$0.009 & 0.792\scriptsize$\pm$0.007 \\
    \textsc{GraSP} (w/o sum)                & 0.666\scriptsize$\pm$0.006 & 0.913\scriptsize$\pm$0.002 & 0.797\scriptsize$\pm$0.003 & \underline{0.735\scriptsize$\pm$0.009} & \underline{0.797\scriptsize$\pm$0.007} \\
    \textsc{GraSP} (w/o NTN)                & 0.662\scriptsize$\pm$0.003 & 0.911\scriptsize$\pm$0.001 & 0.793\scriptsize$\pm$0.002 & 0.718\scriptsize$\pm$0.004 & 0.782\scriptsize$\pm$0.005 \\
    \textsc{GraSP} & \textbf{0.640\scriptsize$\pm$0.013} & \textbf{0.917\scriptsize$\pm$0.002} & \textbf{0.804\scriptsize$\pm$0.003} & \textbf{0.741\scriptsize$\pm$0.009} & \textbf{0.800\scriptsize$\pm$0.002} \\ \bottomrule
    \end{tabular}}
    \vspace{-2mm}
    \caption{Ablation study on AIDS700nef under GED metric.}\label{table:5.4}
    \vspace{-3mm}
\end{table}

\noindent\textbf{Ablation study.} We first compare \textsc{GraSP} with RGGC backbone over \textsc{GraSP} with GCN and GIN backbones.
We also ablate the positional encoding, the attention pooling and summation polling in the multi-scale pooling, and the NTN, denoted as w/o pe, w/o att, w/o sum, and w/o NTN respectively.
The results on AIDS700nef for GED are reported in Table \ref{table:5.4}.
Observe that \textsc{GraSP} obtains the best performance on all metrics than all its ablated versions, which proves the effectiveness of all our proposed components in \textsc{GraSP}.
In particular, with only either attention or summation pooling, the performance is inferior to \textsc{GraSP} with the proposed multi-scale pooling technique that hybrids both pooling techniques, which validates the rationale for designing the technique.
Similar results for the other three datasets are presented in 
the Appendix.

\subsection{Efficiency} \label{section: 5.3}

\begin{figure}[h!]
    \centering
    \vspace{-2mm}
    \includegraphics[width=0.9\columnwidth]{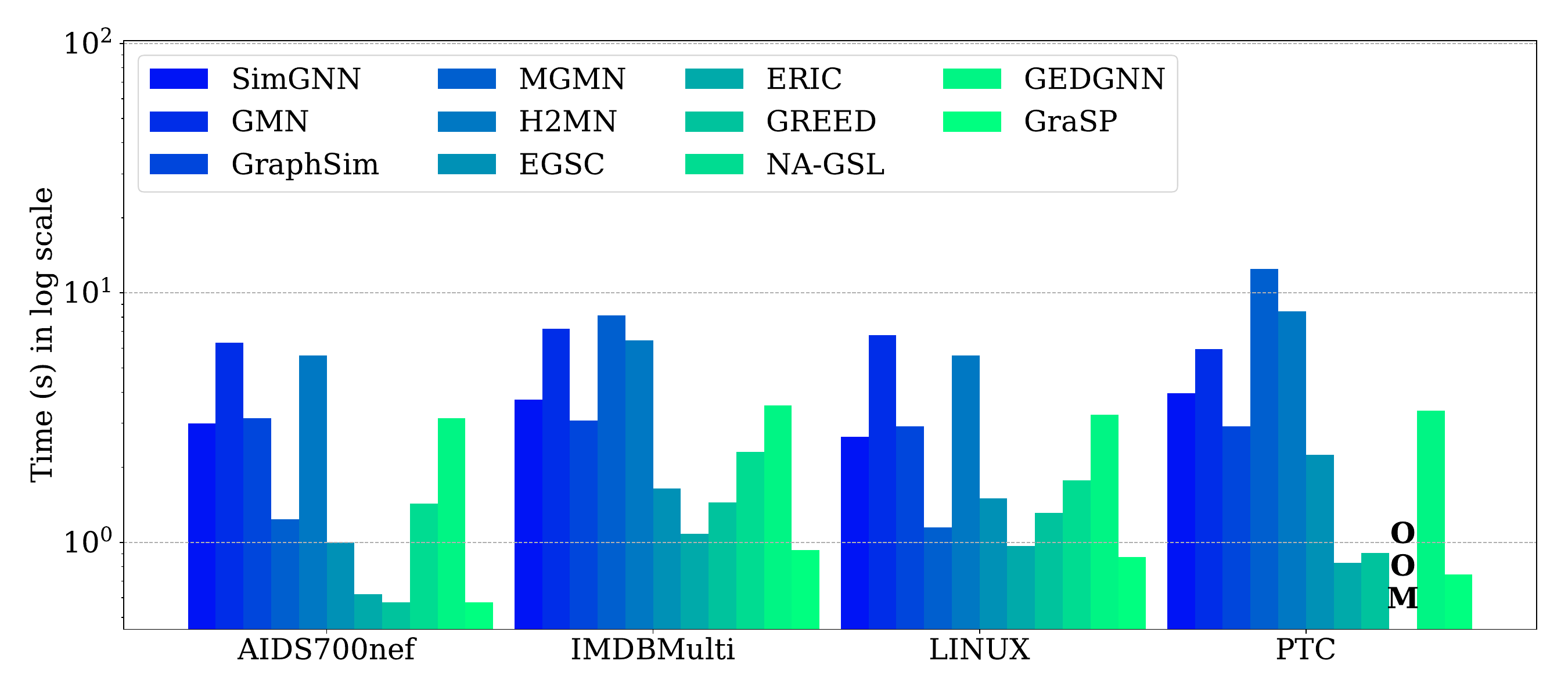}
        \vspace{-3mm}
        \caption{Inference time in second(s) per 10k pairs.}
        \label{fig: 5.3}
        \vspace{-3mm}
\end{figure}

We report the inference time in log scale per 10k graph pairs of all approaches on every dataset in Figure \ref{fig: 5.3}. We omit the results when a certain approach is OOM.
Our method \textsc{GraSP} is the fastest method on all datasets to complete the inference. 
The efficiency of \textsc{GraSP} is due to the following designs. First, \textsc{GraSP} does not need the expensive cross-graph node-level interactions that are usually adopted in existing methods. Second, the positional encoding used in \textsc{GraSP} to enhance node features can be precomputed and reused for the efficiency of online inference. Third, all the technical components in \textsc{GraSP} are designed to make the complicated simple and effective for graph similarity predictions, resulting in efficient performance.
The inference time of \textsc{SimGNN}, GMN, \textsc{GraphSim}, MGMN, NA-GSL, and GEDGNN is longer, which is due to the expensive cross-graph node-level interactions that these models explicitly perform during inference. EGSC, ERIC, and \textsc{Greed} are relatively faster but do not exceed our method.

\begin{figure*}[h!] \centering
  \includegraphics[width=0.92\textwidth]{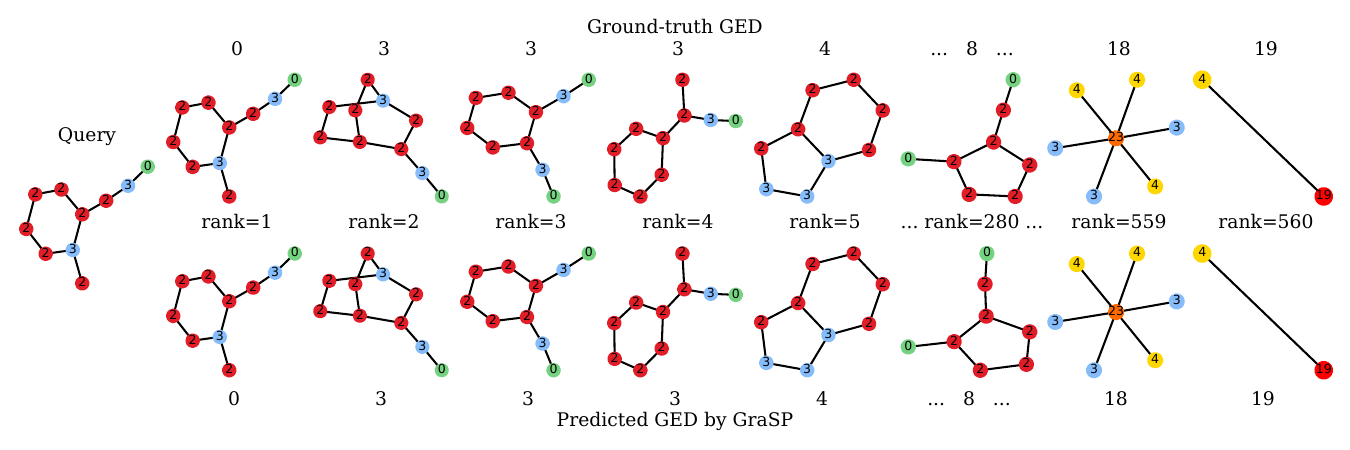}
  \vspace{-4mm}
  \caption{A ranking case study of GED prediction on AIDS700nef.}
  \label{fig: 5.5}
  \vspace{-3mm}
\end{figure*}

\subsection{Case Study} \label{section: 5.5}

We conduct a case study of   \textsc{GraSP} over drug discovery, where graph similarity search often has an essential role
\cite{ranu_probabilistic_2011}. 
We consider the ranking of returned graphs w.r.t. a query graph. 
Figure \ref{fig: 5.5},
shows a case of ranking under the GED metric on the AIDS700nef dataset.
Given a query graph on the left side of Figure \ref{fig: 5.5}, the first row shows the ground-truth ranking of graphs with GED to the query, and the second row shows the graphs ranked by predicted GED from \textsc{GraSP}.
As shown, \textsc{GraSP} can accurately predict the GED values and rank the graphs with similar structures to the top, and the ranking from 1 to 560 is almost the same as the ground truth.
We provide the case studies on the other three datasets in 
the Appendix.

\section{Related Work} \label{section: 6}
For GED, recent works \cite{kim_inves_2019, kim_nass_2020,chang_speeding_2020} compress the search space for faster filtering and verification, but still remain inefficient. For MCS, recent work \cite{mccreesh_partitioning_2017} introduces a branch and bound algorithm that compresses the memory and computational requirements of the search process. Approximation methods like Beam \cite{neuhaus_fast_2006}, Hungarian \cite{riesen_approximate_2009} and VJ \cite{fankhauser_speeding_2011} employ heuristic search and trade precision for reduced complexity, but still have sub-exponential or cubical cost.   \cite{wang_combinatorial_2021} uses a combinatorial technique to combine heuristic and learning methods, but still does not scale well \cite{ranjan_greed_2023}.

In recent years, many learning-based methods have emerged which achieve accurate prediction of similarity values between graphs, while allowing fast predictions. \textsc{SimGNN} \cite{bai_simgnn_2020} is a pioneer work that adopts a Neural Tensor Network (NTN) to capture the graph-level interaction information, and also uses histogram features of node embeddings to capture fine-grained node-level interaction information. GMN \cite{li_graph_2019} encodes cross-graph node matching similarities into node embeddings.%
\textsc{GraphSim} \cite{bai_learning-based_2020} directly captures multi-scale node-level interactions via node embeddings, thereby eliminating the need for computing graph-level embeddings. MGMN \cite{ling_multilevel_2021} introduces a node-graph matching layer when obtaining graph embeddings to capture interactions across levels, i.e., between nodes and graphs. H2MN \cite{zhang_h2mn_2021} transforms the raw graphs into hypergraphs and uses subgraphs pooled from hyperedges of the hypergraphs to conduct subgraph matching. ERIC \cite{zhuo_efficient_2022} proposes a soft matching module to be used during training while to be removed when inference to speed up inference time. NA-GSL \cite{tan_exploring_2023} designs a self-attention module to obtain node embeddings and employs a cross-graph attention module to obtain graph similarity matrices. GEDGNN \cite{piao_computing_2023} adopts a cross-graph node-matching with a graph-matching loss, and conducts a post-processing procedure to generate a graph edit path. These methods explicitly use the cross-graph node-level interactions, thus causing quadratic time costs.
However, the cross-graph node-level interaction module may not be necessary, and simple but effective designs can already achieve superior performance. EGSC \cite{qin_slow_2021} first proposes to omit this cross-graph node-level interaction module. It further uses knowledge distillation to extract the knowledge learned by the larger teacher model to get a lighter-weight student model. Though faster inference times are guaranteed, the student model would perform worse than the teacher model especially when inferring unseen graph queries with different data queries. %
\textsc{Greed} \cite{ranjan_greed_2023} %
further proposes the concept of graph pair-independent embeddings. %
Nevertheless, as shown in experiments, there is still improvement room in its efficacy.
Our model includes a novel embedding structure that incorporates a positional encoding technique to promote the expressiveness power of the node and graph embeddings over the 1-WL test, as well as a new multi-scale pooling technique, to improve performance.

\section{Conclusion}
We present \textsc{GraSP}, a simple but effective method for accurate predictions on GED and MCS.
To make the complicated simple, we design a series of rational and effective techniques in 
\textsc{GraSP} to achieve superior performance.   
We propose techniques to enhance node features via positional encoding, employ a robust graph neural network, and develop a multi-scale pooling technique.
We theoretically prove that \textsc{GraSP} is expressive and passes the 1-WL test, with efficient time complexity linear to the size of a graph pair. 
In extensive experiments, \textsc{GraSP} is versatile in predicting GED and MCS scores accurately on real-world data, often outperforming existing methods by a significant margin.

\section{Acknowledgments}
The work described in this paper was supported by grants from the Research Grants Council of Hong Kong Special Administrative Region, China (No. PolyU 25201221, PolyU 15205224).
Jieming Shi is supported by NSFC No. 62202404, Otto Poon Charitable Foundation Smart Cities Research Institute (SCRI) P0051036-P0050643.
This work is supported by
Tencent Technology Co., Ltd. P0048511.
Renchi Yang is supported by the NSFC Young Scientists Fund (No. 62302414) and the Hong Kong RGC ECS grant (No. 22202623).

\bibliography{aaai25}

\clearpage
\appendix
\section{Proofs}
\subsection{Proof of Proposition \ref{prop: power}} \label{app: A.1}
\begin{proof}
    Suppose that the 1-WL test still fails to distinguish between $\mathcal{G}_1$ and $\mathcal{G}_2$ through $n$ iterations. This is equivalent to the fact that the set of node representations $\{\mathbf{h}_u^\ell\}$ generated is the same for any layer $\ell$ from the 1-st to the n-th layer of the standard MPGNN. When the generated set of node representations is the same, regardless of what pooling techniques are used, the final graph representations are also the same.
    
    At $\ell=0$, it is clear that the set of node label features of two graphs $\{\boldsymbol{\mu}_u\}$ is the same, and since the two feature sets $\{\mathbf{p}_u|u \in \mathcal{G}_1 \}$ and $\{\mathbf{p}_v |v \in \mathcal{G}_2 \}$ are not the same, according to Eq.~\eqref{eq. 2}, where positional coding has been used to enhance the initial node representations, the sets of node representations at $\ell=0$ of the two graphs, $\{\mathbf{h}_u^{(0)}\}=\{\textrm{CONCAT}(\boldsymbol{\mu}_u, \mathbf{p}_u)\}$ and $\{\mathbf{h}_v^{(0)}\}$ are not the same. Therefore, due to the usage of node concatenated representations in Eq.~\eqref{eq. 4}, the sets of concatenated representations $\{\mathbf{h_u}\}$ and $\{\mathbf{h_v}\}$ is not the same. Therefore, the final two graph embeddings are different.
\end{proof}

\section{Descriptions of datasets}
The detailed statics of the dataset are listed in Table \ref{table: A.2}.
AIDS\footnote{\url{https://wiki.nci.nih.gov/display/NCIDTPdata/AIDS+Antiviral+Screen+Data}.} dataset consists of compounds that exhibit anti-HIV properties after screening. A total of 700 compounds of which less than or equal to 10 nodes were selected by \cite{bai_simgnn_2020} to form the AIDS700nef dataset. There are 29 node labels in the AIDS700nef. The IMDBMulti \cite{cao_deep_2015} is a movie collaboration dataset, where nodes represent an actor and edges indicate whether two actors appear in the same movie.  The LINUX dataset consists of a series of Program Dependency Graphs (PDGs) generated by \cite{wang_efficient_2012}, where a node denotes a statement and an edge denotes a dependency between two statements. The LINUX dataset we used in our experiments consists of 1000 graphs randomly selected by \cite{bai_simgnn_2020} in the original LINUX dataset.  The PTC \cite{toivonen_statistical_2003} dataset contains a series of compounds labeled according to their carcinogenicity in male and female mice and rats. There are 19 node labels in the PTC.

\begin{table}[ht] \small \centering
\resizebox{0.9\columnwidth}{!}{
\begin{tabular}{@{}lcccc@{}}
\toprule
Datasets   & \# Graphs & \# Pairs & \# Features & Avg \# Nodes \\ \hline
AIDS700nef & 700                & 78400             & 29                   & 8.9                   \\
IMDBMulti  & 1500               & 360000            & 1                    & 13.0                   \\
LINUX     & 1000               & 160000            & 1                    & 7.6                  \\
PTC        & 344                & 18975             & 19                   & 25.6                   \\ \bottomrule            

\end{tabular}
}
\caption{\label{table: A.2}Statistics of datasets.}
\end{table}

\section{Additional Experiments}
\subsection{Sensitivity Analysis and Hyperparameter Settings} \label{app: A.3}

\begin{figure}[!t] \centering
     \begin{subfigure}[b]{0.32\columnwidth}
         \centering
         \includegraphics[width=\textwidth]{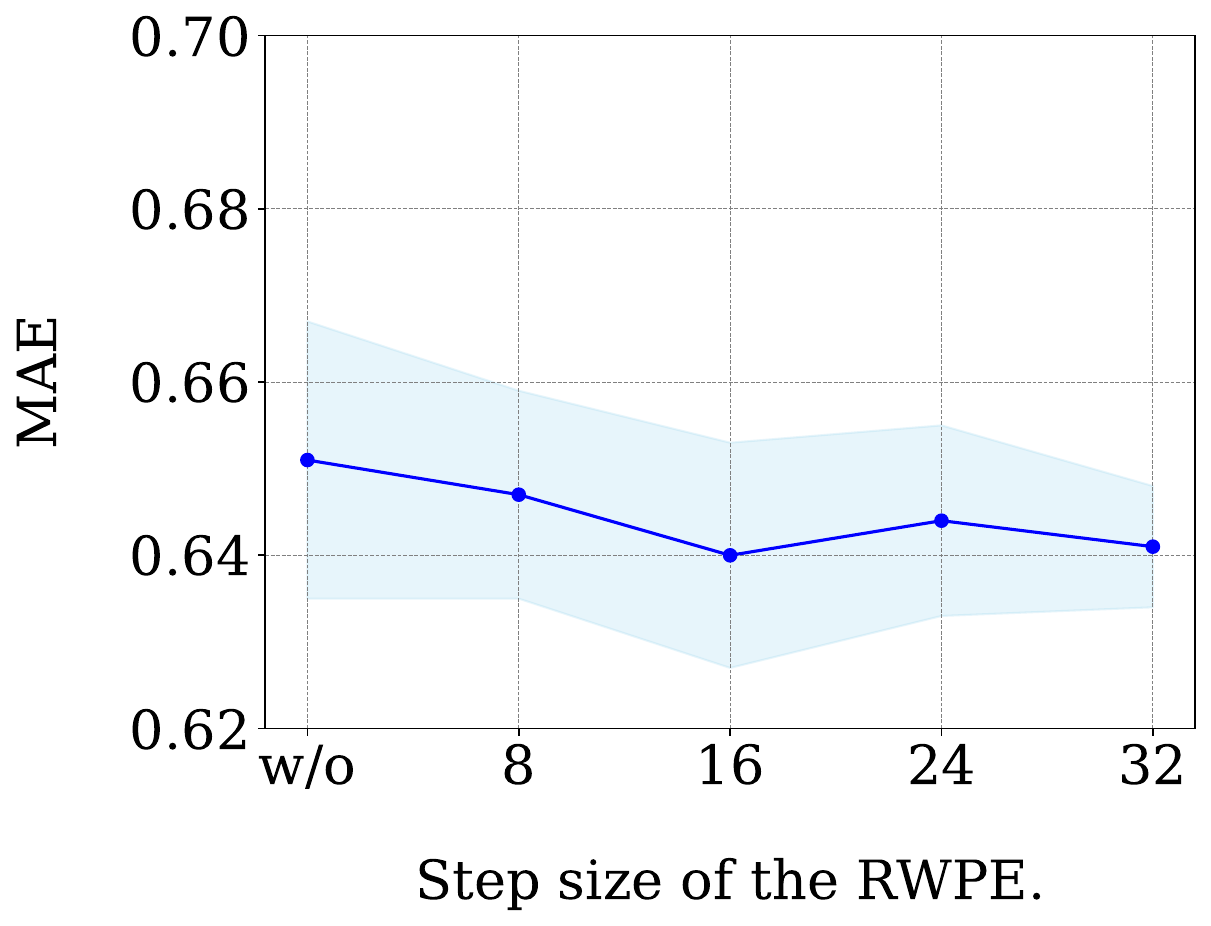}
     \end{subfigure}
     \hfill
     \begin{subfigure}[b]{0.32\columnwidth}
         \centering
         \includegraphics[width=\textwidth]{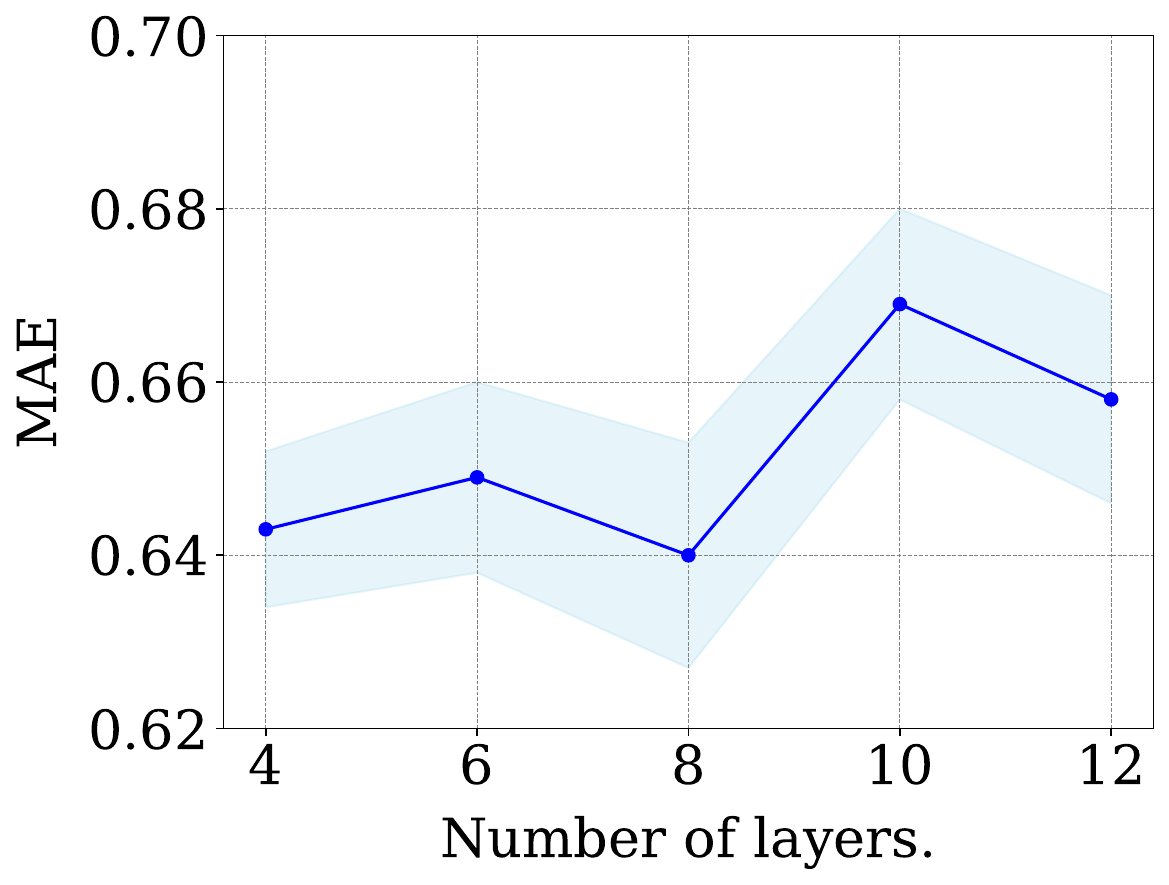}
     \end{subfigure}
     \hfill
     \begin{subfigure}[b]{0.32\columnwidth}
         \centering
         \includegraphics[width=\textwidth]{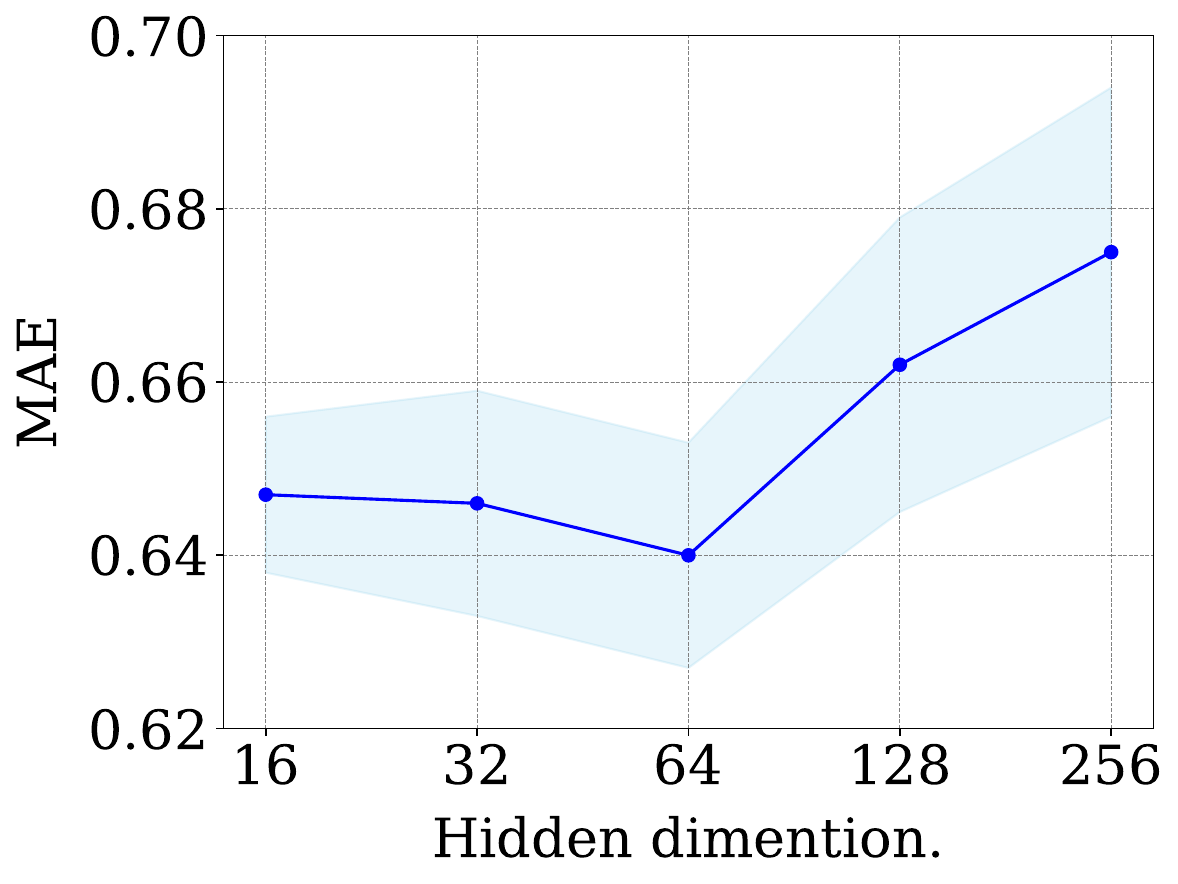}
     \end{subfigure}
    \caption{Hyperparameters sensitivity analysis.}
    \label{fig: A.4}
\end{figure}

We evaluate how the step size $k$ of the RWPE, the number of layers $\ell$ of the GNN backbone, and the hidden dimension $d$ of the node representations will affect the performance of the AIDS700nef dataset. We list the MAE values of the AIDS700nef in Figure \ref{fig: A.4}. We find that the model achieves optimal performance with our hyperparameter settings $k=16$, $\ell=8$, and $d=64$. We can also observe that the use of RWPE improves the performance first and then stabilizes after $k>16$. This is due to the fact that the average number of nodes in AIDS700nef is less than 10, and also that the properties of RWPE that we utilize ensure that when $k$ is large enough, it can be guaranteed that the two non-isomorphic graphs have different sets of RWPEs. When $\ell$ and $d$ are too large, the performance will drop because of overfitting.

The hyperparameter settings are listed in Table \ref{table: A.3}.

\begin{table}[!t] \small \centering
\setlength\tabcolsep{3pt} 
\resizebox{\columnwidth}{!}{\begin{tabular}{@{}clcccc@{}}
\toprule
\multicolumn{2}{c}{Params}           & AIDS700nef             & IMDBMulti              & LINUX                  & PTC                    \\ \midrule
\multirow{7}{*}{GED} & learning rate & $1e^{-4}$ to $1e^{-3}$ & $1e^{-4}$ to $1e^{-3}$ & $2e^{-4}$ to $2e^{-3}$ & $1e^{-4}$ to $1e^{-3}$ \\
                     & weight decay  & $5e^{-4}$              & $5e^{-4}$              & $5e^{-4}$              & $5e^{-4}$              \\
                     & batch size    & 256                    & 256                    & 256                    & 256                     \\
                     & epochs        & $3e^4$                 & $3e^4$                 & $2e^4$                 & $5e^3$                 \\
                     & \# of gnn layers    & 8                      & 4                      & 8                      & 8                      \\
                     & hidden dims.  & 64                     & 64                     & 64                     & 64                     \\
                     & RWPE dims.    & 16                     & 16                     & 10                     & 20                     \\ \midrule
\multirow{7}{*}{MCS} & learning rate & $1e^{-4}$ to $1e^{-3}$ & $1e^{-4}$ to $1e^{-3}$ & $2e^{-4}$ to $2e^{-3}$ & $1e^{-4}$ to $1e^{-3}$ \\
                     & weight decay  & $5e^{-4}$              & $5e^{-4}$              & $5e^{-4}$              & $5e^{-4}$              \\
                     & batch size    & 256                    & 256                    & 256                    & 256                     \\
                     & epochs        & $3e^4$                 & $3e^4$                 & $2e^4$                 & $2e^4$                 \\
                     & \# of gnn layers    & 8                      & 4                      & 8                      & 8                      \\
                     & hidden dims.  & 64                     & 64                     & 64                     & 64                     \\
                     & RWPE dims.    & 16                     & 16                     & 10                     & 20                     \\ \bottomrule
\end{tabular}}
\caption{\label{table: A.3} Hyperparamater settings on 4 datasets.}
\end{table}

\subsection{A comparison on GED prediction error heatmap} \label{app: A.6}
\begin{figure}[!t] \centering
     \begin{subfigure}[b]{0.32\columnwidth}
         \centering
         \includegraphics[height=2cm]{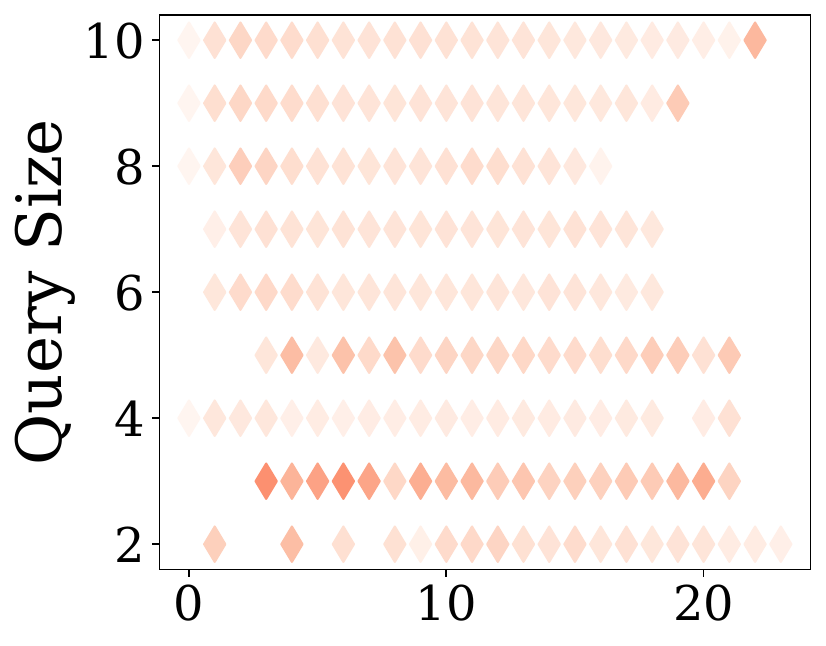}
         \caption{\textsc{GraSP}}
         \label{fig: A.6.1}
     \end{subfigure}
     \hfill
     \begin{subfigure}[b]{0.32\columnwidth}
         \centering
         \includegraphics[height=2cm]{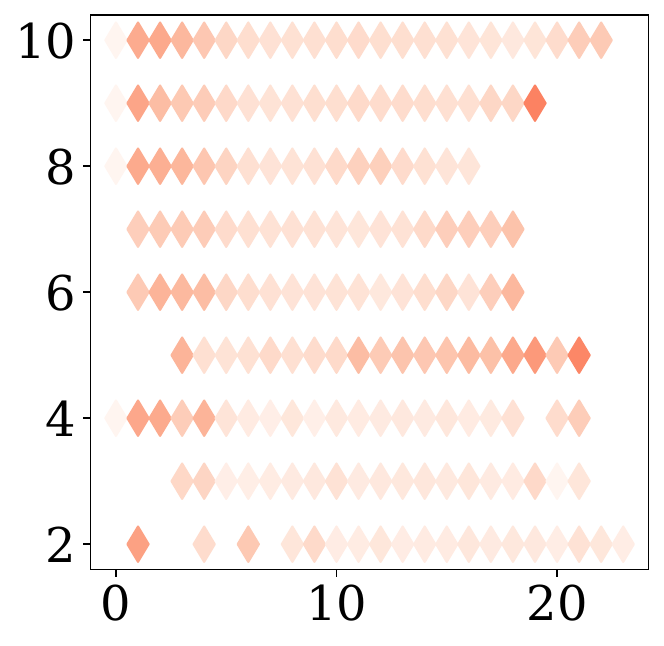}
         \caption{\textsc{Greed}}
         \label{fig: A.6.2}
     \end{subfigure}
     \hfill
     \begin{subfigure}[b]{0.32\columnwidth}
         \centering
         \includegraphics[height=2.052cm]{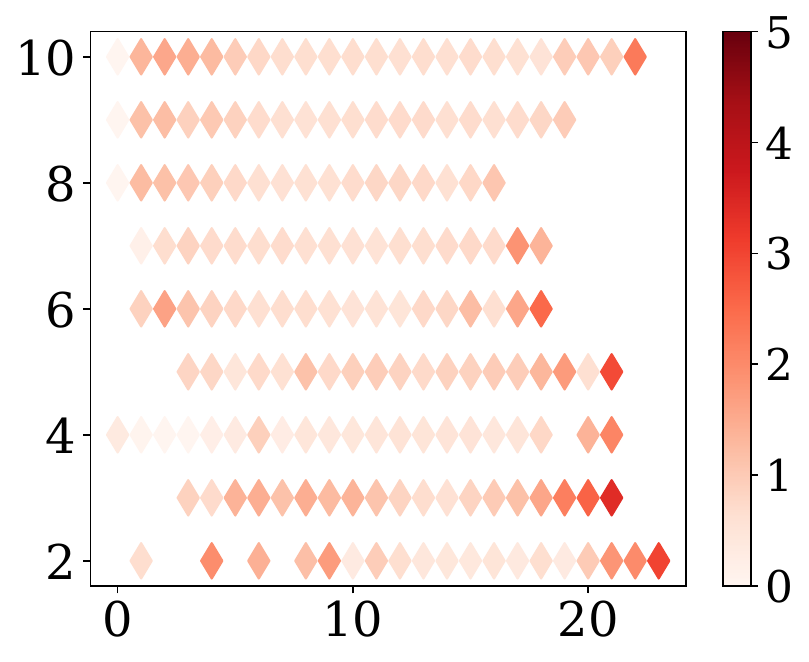}
         \caption{ERIC}
         \label{fig: A.6.3}
     \end{subfigure}
    \caption{Absolute error heatmap on GED on AIDS700nef.}
    \label{fig: A.6}
\end{figure}

\begin{figure}[!t] \centering
     \begin{subfigure}[b]{0.32\columnwidth}
         \centering
         \includegraphics[height=2cm]{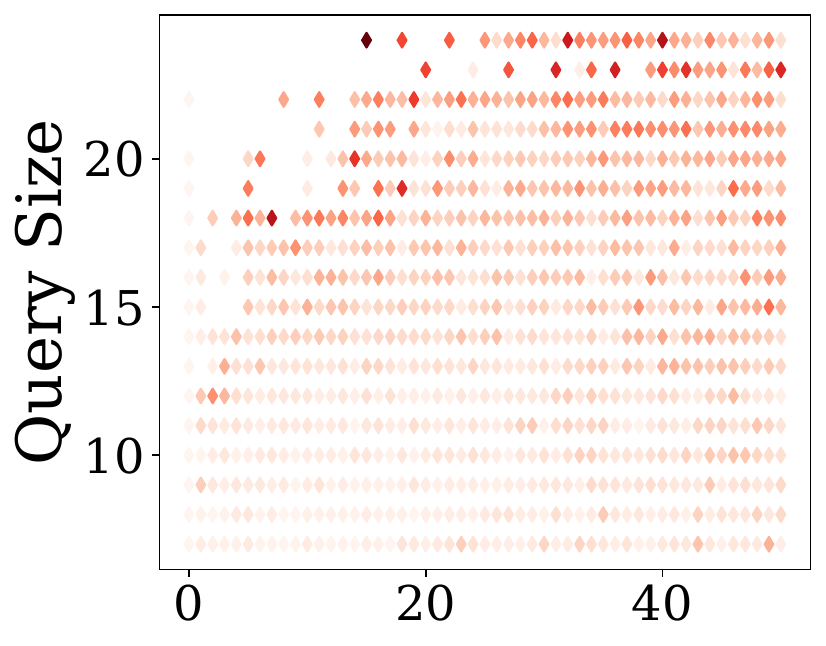}
         \caption{\textsc{GraSP}}
         \label{fig: A.6-imdb.1}
     \end{subfigure}
     \hfill
     \begin{subfigure}[b]{0.32\columnwidth}
         \centering
         \includegraphics[height=2cm]{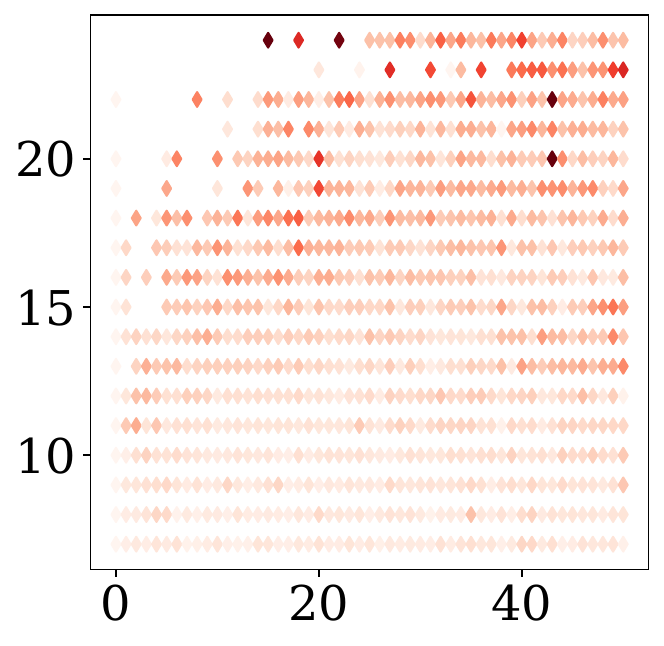}
         \caption{\textsc{Greed}}
         \label{fig: A.6-imdb.2}
     \end{subfigure}
     \hfill
     \begin{subfigure}[b]{0.32\columnwidth}
         \centering
         \includegraphics[height=2.052cm]{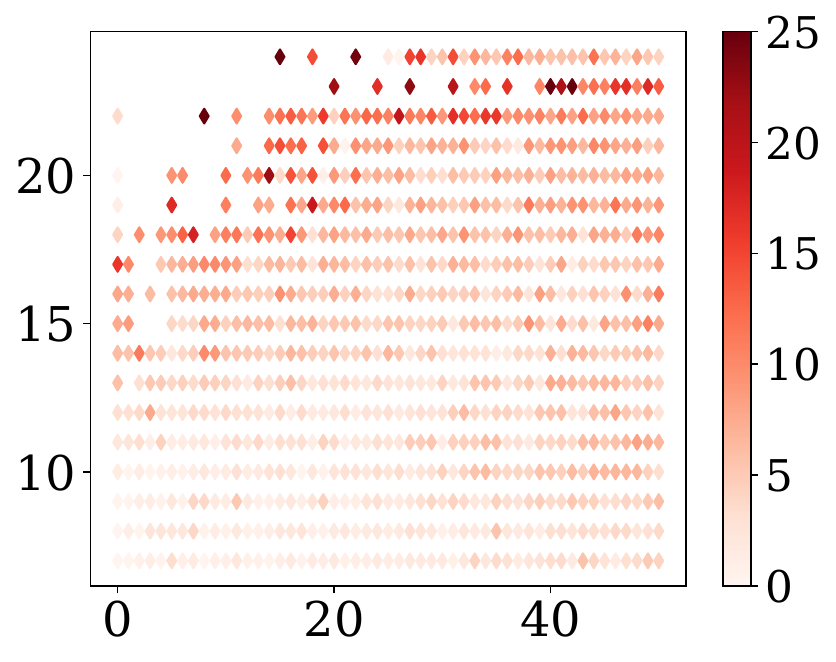}
         \caption{ERIC}
         \label{fig: A.6-imdb.3}
     \end{subfigure}
    \caption{Absolute error heatmap on GED on IMDBMulti.}
    \label{fig: A.6-imdb}
\end{figure}

\begin{figure}[!t] \centering
     \begin{subfigure}[b]{0.32\columnwidth}
         \centering
         \includegraphics[height=2cm]{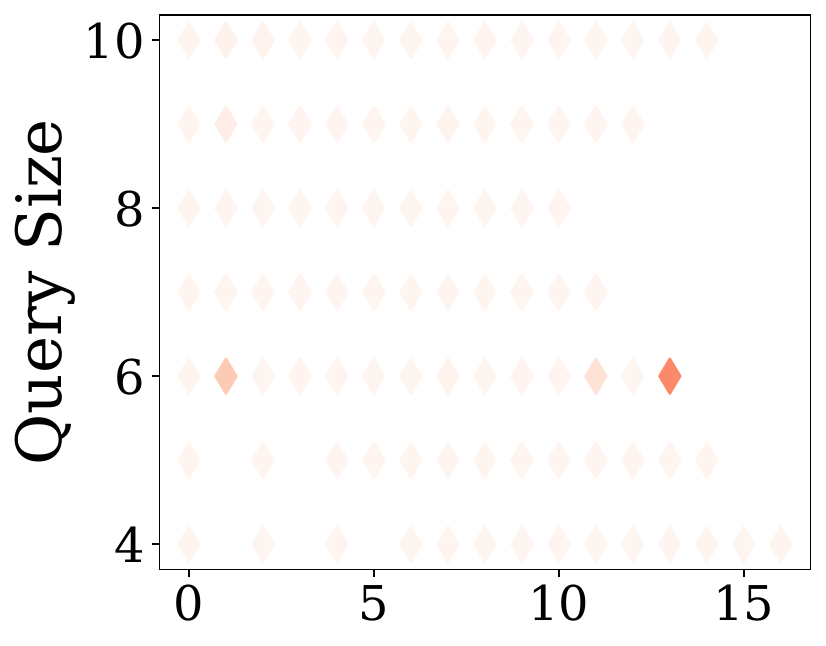}
         \caption{\textsc{GraSP}}
         \label{fig: A.6-linux.1}
     \end{subfigure}
     \hfill
     \begin{subfigure}[b]{0.32\columnwidth}
         \centering
         \includegraphics[height=2cm]{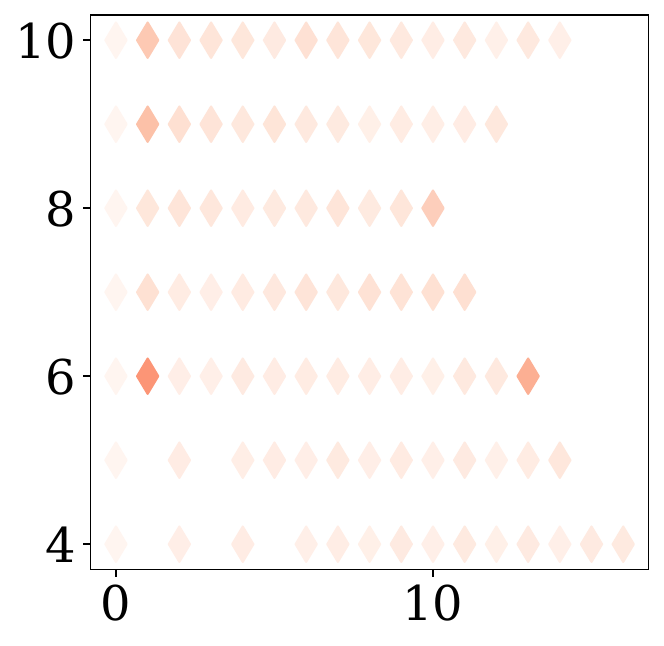}
         \caption{\textsc{Greed}}
         \label{fig: A.6-linux.2}
     \end{subfigure}
     \hfill
     \begin{subfigure}[b]{0.32\columnwidth}
         \centering
         \includegraphics[height=2.052cm]{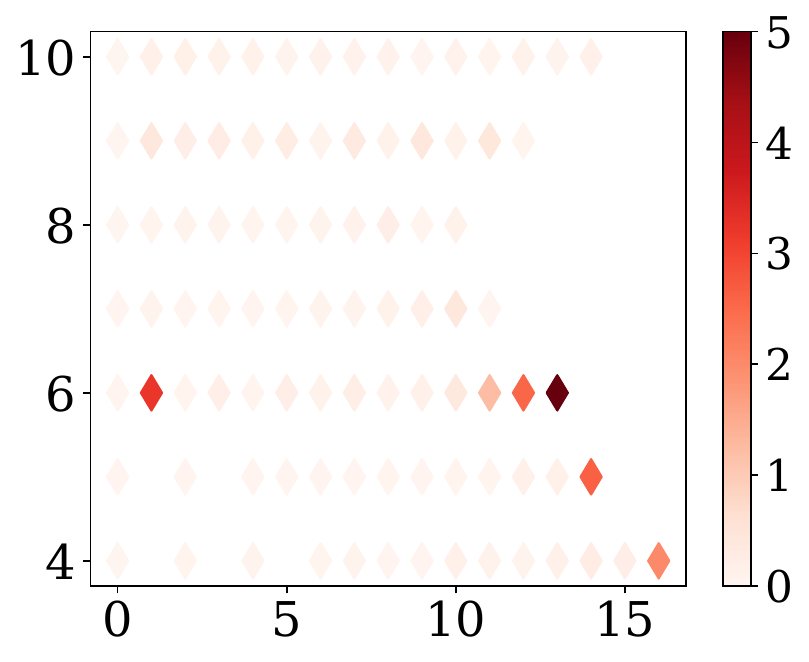}
         \caption{ERIC}
         \label{fig: A.6-linux.3}
     \end{subfigure}
    \caption{Absolute error heatmap on GED on LINUX.}
    \label{fig: A.6-linux}
\end{figure}

\begin{figure}[!t] \centering
     \begin{subfigure}[b]{0.32\columnwidth}
         \centering
         \includegraphics[height=2cm]{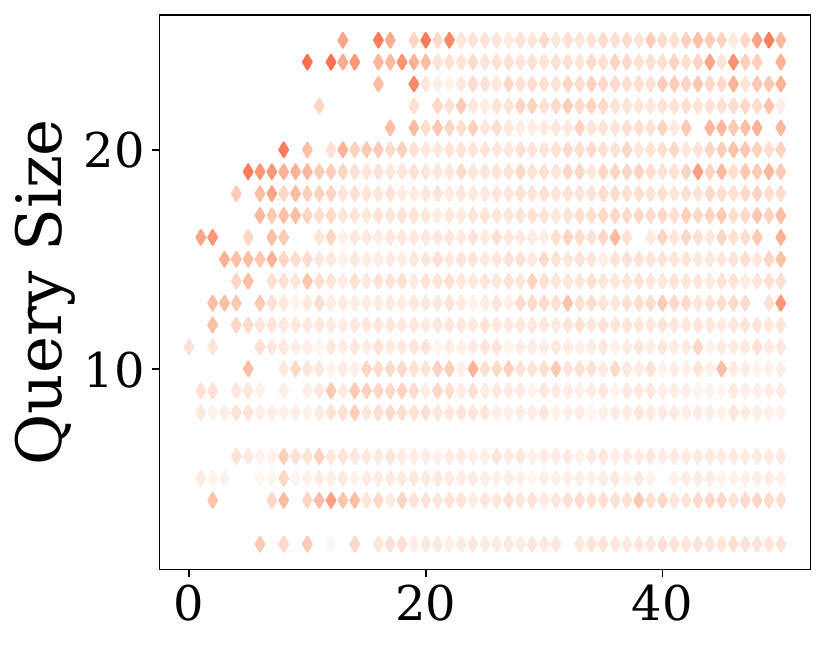}
         \caption{\textsc{GraSP}}
         \label{fig: A.6-ptc.1}
     \end{subfigure}
     \hfill
     \begin{subfigure}[b]{0.32\columnwidth}
         \centering
         \includegraphics[height=2cm]{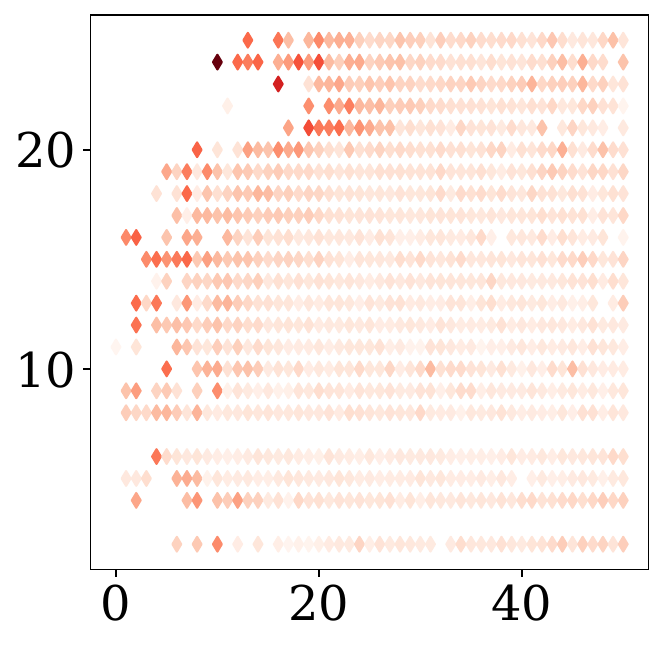}
         \caption{\textsc{Greed}}
         \label{fig: A.6-ptc.2}
     \end{subfigure}
     \hfill
     \begin{subfigure}[b]{0.32\columnwidth}
         \centering
         \includegraphics[height=2.052cm]{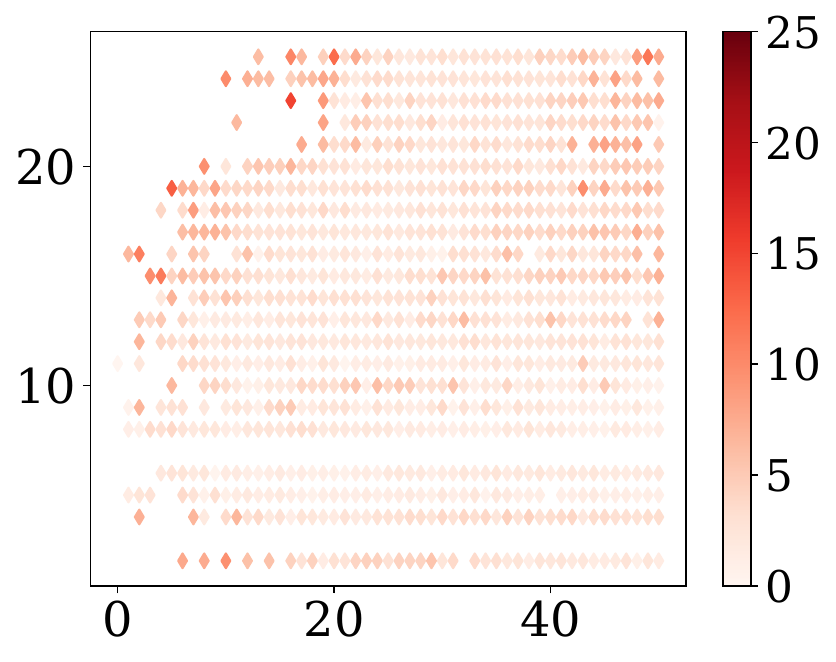}
         \caption{ERIC}
         \label{fig: A.6-ptc.3}
     \end{subfigure}
    \caption{Absolute error heatmap on GED on PTC.}
    \label{fig: A.6-ptc}
\end{figure}

The absolute GED error heatmaps of our approach, \textsc{Greed} and ERIC on four datasets are shown in Figure \ref{fig: A.6}, \ref{fig: A.6-imdb}, \ref{fig: A.6-linux} and \ref{fig: A.6-ptc}. The x-axis represents the GED between the query graph and the target graphs. The y-axis represents the number of nodes in the query graph. The color of each dot represents the absolute error on GED between a query graph and a target graph. The lighter color indicates a lower absolute error. 
Our method \textsc{GraSP} has better performance than existing methods over different query sizes and GED values on all datasets.

\subsection{Visualization.} 
\begin{figure}[!t] \centering
     \begin{subfigure}[b]{0.32\columnwidth}
         \centering
         \frame{\includegraphics[width=\textwidth]{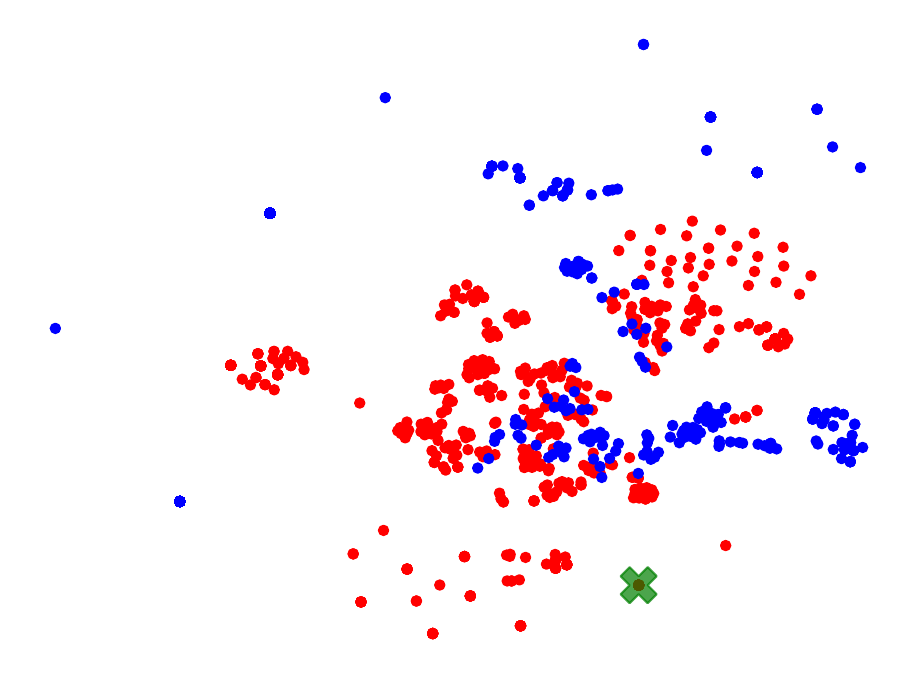}}
         \caption{Using attention.}
         \label{fig: A.7.1}
     \end{subfigure}
     \hfill
     \begin{subfigure}[b]{0.32\columnwidth}
         \centering
         \frame{\includegraphics[width=\textwidth]{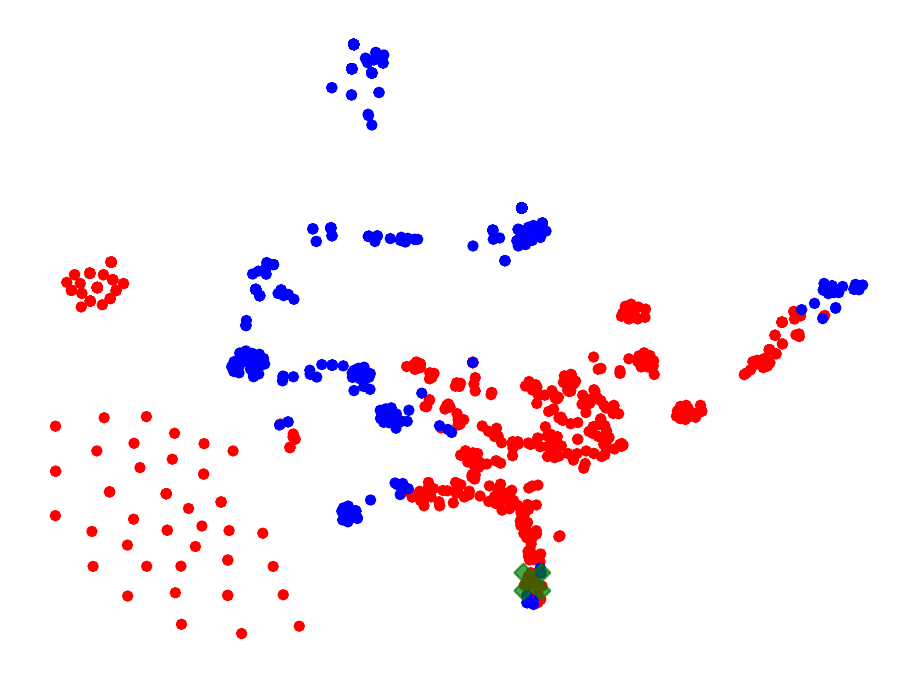}}
         \caption{Using add.}
         \label{fig: A.7.2}
     \end{subfigure}
     \hfill
     \begin{subfigure}[b]{0.32\columnwidth}
         \centering
         \frame{\includegraphics[width=\textwidth]{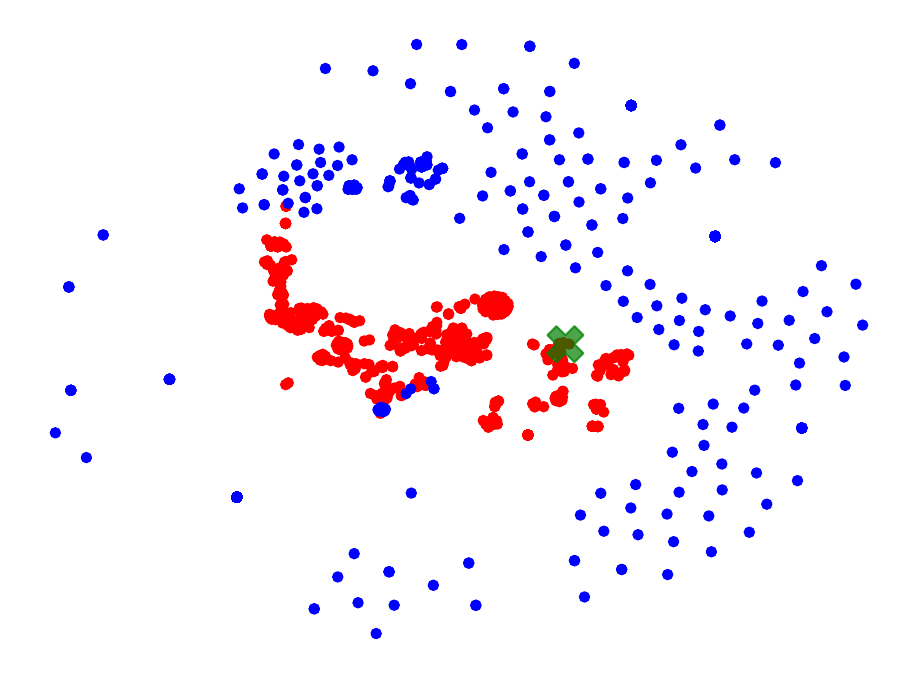}}
         \caption{Using multi.}
         \label{fig: A.7.3}
     \end{subfigure}
    \caption{T-SNE visualization on IMDBMulti with attention pooling, add pooling, and multi-scale pooling. We plot the graph embeddings on a 2d plain, where the green cross denotes the query graph, red dots denote the top $50\%$ of similar graphs in the database and blue dots denote the remaining graphs.}
    \label{fig: A.7}
\end{figure}

To exemplify the effect of multi-scale pooling, we conducted experiments on the IMDBMulti dataset with t-SNE visualization \cite{maaten_visualize_2008}, as shown in Figure \ref{fig: A.7}. 
Compared to the graph embeddings obtained using only attention and summation pooling methods, the graph embeddings obtained using the multi-scale pooling technique exhibit better patterns in the embedding space, which reflects the effective modeling of graph similarity properties.

\subsection{Generalization.}
We follow the setting in \cite{ranjan_greed_2023} to demonstrate the generalization power of \textsc{GraSP} on large unseen graphs for GED predictions.
Specifically, we get \textsc{GraSP}-25 and \textsc{GraSP}-50 by training \textsc{GraSP} on graphs with node sizes up to 25 and 50, respectively, and then test on graph pairs in which the number of nodes in the query graph is in the range [25, 50]. Table \ref{tab:generalizationexp} shows the results, with comparison to  \textsc{Greed} and H2MN. Observe that (i) the performance of all methods degrades for large query sizes in [25,50], compared with the entire set in [0,50], (ii) when trained using smaller graphs from 50 to 25 size, all methods also degrade, (iii) \textsc{GraSP}-25 (resp. \textsc{GraSP}-50) keeps the best performance than the baselines under all these generalization settings.

\begin{table}[!t] \small \centering
\resizebox{0.7\columnwidth}{!}{
\begin{tabular}{lcc}
\hline
 & \multicolumn{1}{l}{Query Size in {[}0,50{]}} & \multicolumn{1}{l}{Query Size in {[}25, 50{]}} \\ \hline
\textsc{GraSP}-50 & \textbf{3.973} & \textbf{5.002} \\
\textsc{Greed}-50 & 4.574 & 5.323 \\
H2MN-50 & 6.813 & 7.520 \\ \hline
\textsc{GraSP}-25 & \textbf{4.692} & \textbf{6.277} \\
\textsc{Greed}-25 & 5.928 & 6.474 \\
H2MN-25 & 7.023 & 8.115 \\ \hline
\end{tabular}
}
\caption{Generalization on large unseen graphs with MAE results to predict GED on PTC.} \label{tab:generalizationexp}
\end{table}

\subsection{Additional Ablation Studies} \label{app: B.4}
The ablation results on IMDBMulti, LINUX, and PTC datasets are reported in Table \ref{table: ablation-imdb}, \ref{table: ablation-linux} and \ref{table: ablation-ptc}. We also include the average ranking of all metrics. The results show that \textsc{GraSP} has a better overall performance than ablated versions on these three datasets.

\begin{table}[!t]
    \small \centering
    \resizebox{\columnwidth}{!}{\begin{tabular}{@{}lcccccc@{}}
    \toprule
                                  & MAE                             & $\rho$                          & $\tau$                          & P@10                            & P@20                            & rank                             \\ \midrule
    \textsc{GraSP} (GIN)                    & 4.374\scriptsize$\pm$0.230 & 0.941\scriptsize$\pm$0.003 & 0.868\scriptsize$\pm$0.006 & 0.861\scriptsize$\pm$0.003 & 0.872\scriptsize$\pm$0.014 & 4.4 \\
    \textsc{GraSP} (GCN)                    & 4.462\scriptsize$\pm$0.189 & 0.942\scriptsize$\pm$0.002 & 0.870\scriptsize$\pm$0.003 & \underline{0.862\scriptsize$\pm$0.004} & 0.868\scriptsize$\pm$0.010 & 4.4 \\
    \textsc{GraSP} (w/o pe)                 & 3.991\scriptsize$\pm$0.060 & \textbf{0.943\scriptsize$\pm$0.004} & \textbf{0.878\scriptsize$\pm$0.006} & 0.858\scriptsize$\pm$0.009 & 0.871\scriptsize$\pm$0.006 & 3.2 \\
    \textsc{GraSP} (w/o att)                & \underline{3.972\scriptsize$\pm$0.129} & 0.942\scriptsize$\pm$0.002 & 0.873\scriptsize$\pm$0.004 & 0.861\scriptsize$\pm$0.007 & 0.870\scriptsize$\pm$0.006 & 3.4 \\
    \textsc{GraSP} (w/o sum)                & 4.244\scriptsize$\pm$0.148 & 0.940\scriptsize$\pm$0.001 & 0.867\scriptsize$\pm$0.001 & 0.859\scriptsize$\pm$0.003 & \underline{0.873\scriptsize$\pm$0.006} & 4.8 \\
    \textsc{GraSP} (w/o NTN)                & 3.978\scriptsize$\pm$0.285 & 0.930\scriptsize$\pm$0.001 & 0.847\scriptsize$\pm$0.002 & 0.854\scriptsize$\pm$0.008 & 0.863\scriptsize$\pm$0.006 & 6.2 \\
    \textsc{GraSP} & \textbf{3.966\scriptsize$\pm$0.064} & \underline{0.942\scriptsize$\pm$0.001} & \underline{0.874\scriptsize$\pm$0.003} & \textbf{0.863\scriptsize$\pm$0.011} & \textbf{0.876\scriptsize$\pm$0.014} & 1.4 \\ \bottomrule
    \end{tabular}}
    \caption{\label{table: ablation-imdb}Ablation study on IMDBMulti under GED metric.}
\end{table}

\begin{table}[!t]
    \small \centering
    \resizebox{\columnwidth}{!}{\begin{tabular}{@{}lcccccc@{}}
    \toprule
                                  & MAE                             & $\rho$                          & $\tau$                          & P@10                            & P@20                            & rank                             \\ \midrule
    \textsc{GraSP} (GIN)                    & 0.067\scriptsize$\pm$0.003 & 0.972\scriptsize$\pm$0.001 & 0.906\scriptsize$\pm$0.001 & 0.980\scriptsize$\pm$0.005 & 0.981\scriptsize$\pm$0.003 & 6.0 \\
    \textsc{GraSP} (GCN)                    & 0.047\scriptsize$\pm$0.008 & 0.973\scriptsize$\pm$0.001 & 0.908\scriptsize$\pm$0.001 & 0.983\scriptsize$\pm$0.002 & 0.986\scriptsize$\pm$0.002 & 3.6 \\
    \textsc{GraSP} (w/o pe)                 & \underline{0.047\scriptsize$\pm$0.004} & 0.973\scriptsize$\pm$0.001 & 0.908\scriptsize$\pm$0.001 & 0.981\scriptsize$\pm$0.001 & 0.987\scriptsize$\pm$0.002 & 3.6 \\
    \textsc{GraSP} (w/o att)                & 0.049\scriptsize$\pm$0.008 & \underline{0.974\scriptsize$\pm$0.001} & \underline{0.909\scriptsize$\pm$0.001} & \textbf{0.985\scriptsize$\pm$0.004} & \underline{0.992\scriptsize$\pm$0.004} & 2.2 \\
    \textsc{GraSP} (w/o sum)                & 0.049\scriptsize$\pm$0.008 & \underline{0.974\scriptsize$\pm$0.001} & 0.908\scriptsize$\pm$0.001 & \underline{0.984\scriptsize$\pm$0.002} & 0.991\scriptsize$\pm$0.002 & 2.8 \\
    \textsc{GraSP} (w/o NTN)                & 0.321\scriptsize$\pm$0.002 & 0.965\scriptsize$\pm$0.001 & 0.885\scriptsize$\pm$0.001 & 0.979\scriptsize$\pm$0.003 & 0.982\scriptsize$\pm$0.004 & 6.6 \\
    \textsc{GraSP} & \textbf{0.042\scriptsize$\pm$0.003} & \textbf{0.975\scriptsize$\pm$0.001} & \textbf{0.922\scriptsize$\pm$0.001} & 0.983\scriptsize$\pm$0.003 & \textbf{0.992\scriptsize$\pm$0.002} & 1.6 \\ \bottomrule
    \end{tabular}}
    \caption{\label{table: ablation-linux}Ablation study on LINUX under GED metric.}
\end{table}

\begin{table}[!t]
    \small \centering
    \resizebox{\columnwidth}{!}{\begin{tabular}{@{}lcccccc@{}}
    \toprule
                                  & MAE                             & $\rho$                          & $\tau$                          & P@10                            & P@20                            & rank                             \\ \midrule
    \textsc{GraSP} (GIN)                    & 3.744\scriptsize$\pm$0.138 & 0.948\scriptsize$\pm$0.001 & 0.855\scriptsize$\pm$0.002 & 0.594\scriptsize$\pm$0.007 & 0.699\scriptsize$\pm$0.005 & 4.6 \\
    \textsc{GraSP} (GCN)                    & 3.645\scriptsize$\pm$0.122 & 0.948\scriptsize$\pm$0.002 & 0.856\scriptsize$\pm$0.003 & 0.595\scriptsize$\pm$0.007 & 0.699\scriptsize$\pm$0.007 & 4.2 \\
    \textsc{GraSP} (w/o pe)                 & 3.802\scriptsize$\pm$0.138 & 0.948\scriptsize$\pm$0.002 & 0.855\scriptsize$\pm$0.002 & 0.568\scriptsize$\pm$0.005 & 0.693\scriptsize$\pm$0.007 & 6.4 \\
    \textsc{GraSP} (w/o att)                & 3.733\scriptsize$\pm$0.130 & 0.949\scriptsize$\pm$0.001 & \underline{0.857\scriptsize$\pm$0.001} & \underline{0.607\scriptsize$\pm$0.010} & \textbf{0.708\scriptsize$\pm$0.003} & 2.8 \\
    \textsc{GraSP} (w/o sum)                & 3.687\scriptsize$\pm$0.159 & 0.948\scriptsize$\pm$0.001 & 0.856\scriptsize$\pm$0.002 & 0.589\scriptsize$\pm$0.014 & 0.703\scriptsize$\pm$0.009 & 3.8 \\
    \textsc{GraSP} (w/o NTN)                & \textbf{3.367\scriptsize$\pm$0.146} & \underline{0.951\scriptsize$\pm$0.002} & 0.855\scriptsize$\pm$0.004 & 0.580\scriptsize$\pm$0.008 & 0.688\scriptsize$\pm$0.005 & 4.6 \\
    \textsc{GraSP} & \underline{3.556\scriptsize$\pm$0.080} & \textbf{0.952\scriptsize$\pm$0.002} & \textbf{0.861\scriptsize$\pm$0.003} & \textbf{0.612\scriptsize$\pm$0.019} & \underline{0.707\scriptsize$\pm$0.010} & 1.4 \\ \bottomrule
    \end{tabular}}
    \caption{\label{table: ablation-ptc}Ablation study on PTC under GED metric.}
\end{table}

\begin{figure*}[!t] \centering
  \includegraphics[width=0.92\textwidth]{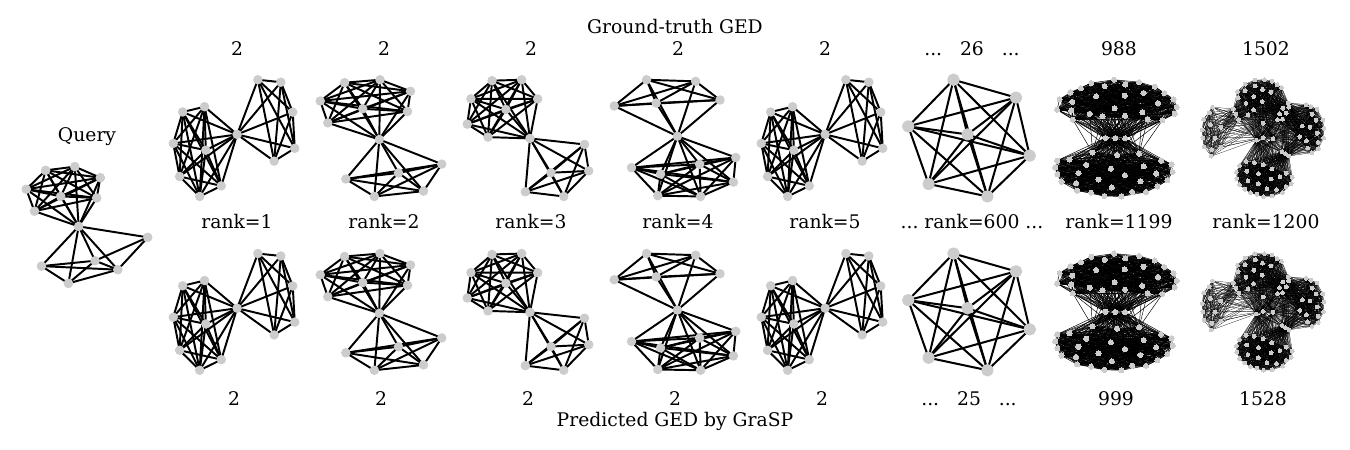}
  \caption{A ranking case on IMDBMulti.}
  \label{fig: A.5.1}
\end{figure*}

\begin{figure*}[!t] \centering
  \includegraphics[width=0.92\textwidth]{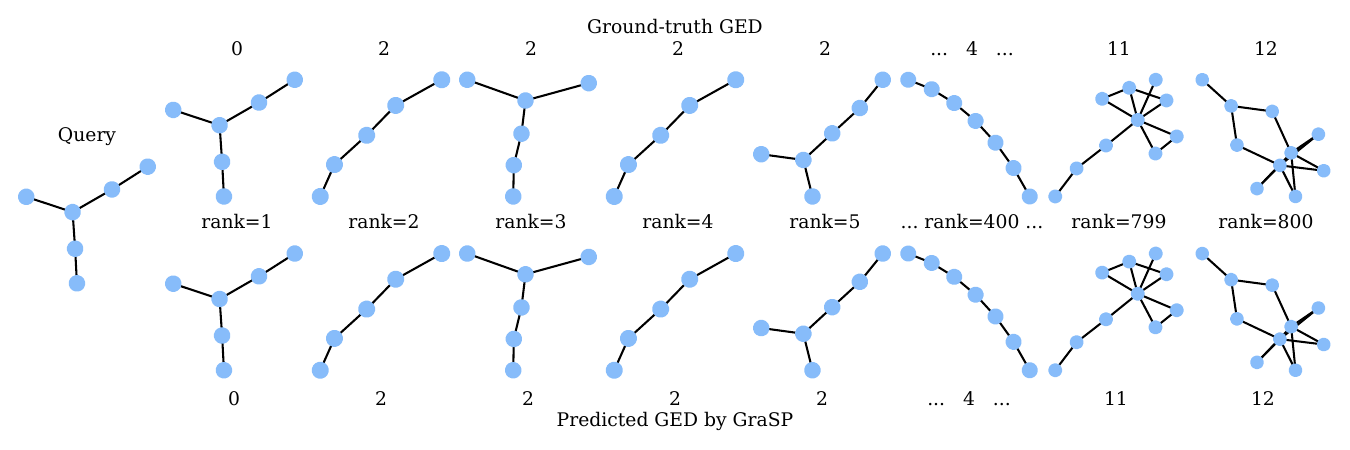}
  \caption{A ranking case on LINUX.}
  \label{fig: A.5.2}
\end{figure*}

\begin{figure*}[!t] \centering
  \includegraphics[width=0.92\textwidth]{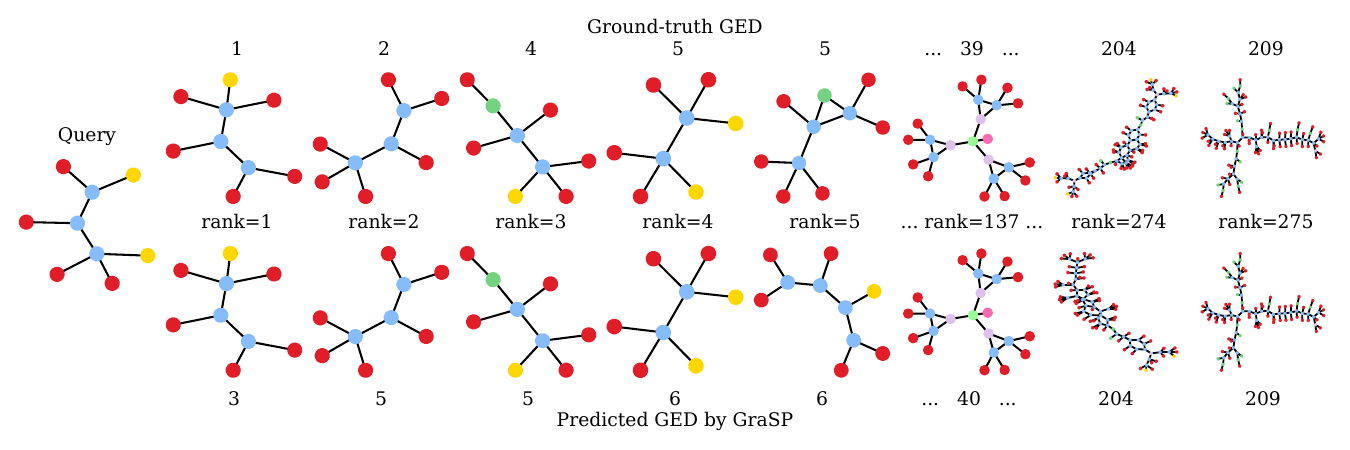}
  \caption{A ranking case on PTC.}
  \label{fig: A.5.3}
\end{figure*}
\subsection{Additional Case Studies} \label{app: A.5}
Three case studies on IMDBMulti, LINUX, and PTC under GED are included in Figure \ref{fig: A.5.1}, \ref{fig: A.5.2} and \ref{fig: A.5.3}. For the AIDS700nef and LINUX datasets, the exact GED values are obtained using the A* algorithm, and the predicted values are generated by our model. However, for the IMDBMulti and PTC datasets, due to the computational infeasibility of obtaining exact GEDs for larger graphs, the ground-truth GEDs are estimated as the minimum values among the Beam, Hungarian, and VJ algorithm outputs.
Including Figure \ref{fig: 5.5}, we observe that for the AIDS700nef, IMDBMulti, and LINUX datasets, the ranking of the predicted GEDs perfectly aligns with that of the ground-truth GEDs. However, for the PTC dataset, discrepancies are observed at rank 4 and rank 274. Nonetheless, our method generally performs well in similarity ranking across datasets of different sizes.

\section{Extension to Consider Edge Relabeling} \label{app: B.3}
Here we discuss how to extend \textsc{GraSP} to include the cost of relabeling edges. Following \cite{dwivedi_benchmarking_2022}, Eq.~\eqref{eq. 3} can be modified as: 
\begin{equation}\label{eq. A. 1}
\small
\mathbf{h}_{i}^{(\ell)} = \mathbf{h}_{i}^{(\ell - 1)} + \textrm{ReLU}\left(\mathbf{W}_{S}\mathbf{h}_{i}^{(\ell - 1)} + \sum _{j \in \mathcal{N}(i)} \mathbf{e}^{(\ell)}_{i,j} \odot \mathbf{W}_{N}\mathbf{h}_{j}^{(\ell - 1)}\right),
\end{equation}
where $\mathbf{e}^{(\ell)}_{i,j}$ is the edge gate and can be defined as
$$
\textstyle
    \mathbf{e}^{(\ell)}_{i,j} = \sigma\left(\mathbf{e}^{(\ell - 1)}_{i,j} + \textrm{ReLU}\left(\mathbf{A}\mathbf{h}_{i}^{(\ell - 1)} + \mathbf{B}\mathbf{h}_{j}^{(\ell - 1)} + \mathbf{C}\mathbf{e}^{(\ell - 1)}_{i,j}\right)\right),
    \label{eq. A. 2}
$$
where $\mathbf{A}, \mathbf{B}, \mathbf{C} \in \mathbb{R} ^ {2d \times 2d}$ and $\mathbf{e}^{(0)}_{i,j}$ represents the input edge feature.

\end{document}